%% file: main.tex
\documentclass[10pt,twocolumn,letterpaper]{article}

\usepackage[pagenumbers]{cvpr} 

\input{preamble}
\usepackage{algorithm,algpseudocode}

\definecolor{cvprblue}{rgb}{0.21,0.49,0.74}
\usepackage[pagebackref,breaklinks,colorlinks,allcolors=cvprblue]{hyperref}

\def\confName{CVPR}
\def\confYear{2027}

\title{M\textsuperscript{2}AdaptVLA: Multimodal Memory-Augmented Vision-Language-Action Model with Predictive Test-Time Adaptation}

\author{Liufan Tan\textsuperscript{1,2,*}, Tong Wang\textsuperscript{1,*}, Yuehao Yin\textsuperscript{1}, Jian Chen\textsuperscript{1,3}, Yihang Yin\textsuperscript{1,4},\\
Ying Qu\textsuperscript{1}, Wenyu Li\textsuperscript{1}, Ruyi Gan\textsuperscript{1}, Hao Wang\textsuperscript{1,\ensuremath{\dagger}}}

\ifdefined\SubmissionOnly\else\def\WithSupplement{}\fi
\newcommand{\suppref}[2]{\ifdefined\WithSupplement\ref{#1}\else#2\fi}

\makeatletter
\patchcmd{\@maketitle}{\vspace*{24pt}}{\vspace*{14pt}}{}{}
\patchcmd{\@maketitle}{\vspace*{12pt}}{\vspace*{2pt}}{}{}
\makeatother

\begin{document}
\maketitle
\input{fig/teaser}
\begingroup
\makeatletter \long\def\@makefntext#1{\noindent#1}\makeatother \renewcommand{\thefootnote}{}
\footnotetext{\textsuperscript{1}X SQUARE ROBOT\quad \textsuperscript{2}Chongqing University\newline
\textsuperscript{3}University of Science and Technology of China\newline
\textsuperscript{4}University of Edinburgh\newline
\textsuperscript{*}Equal contribution.\quad \textsuperscript{\ensuremath{\dagger}}Corresponding author.}
\endgroup
\input{sec/0_abstract}    
\input{sec/1_intro}
\input{sec/2_formatting}
\input{sec/3_finalcopy}
{
    \small
    \bibliographystyle{ieeenat_fullname}
    \bibliography{main}
}

\ifdefined\WithSupplement
  \input{sec/X_suppl_from_original}
\fi

\end{document}

%% file: preamble.tex
\usepackage{cuted} 

\usepackage{currfile} 

\usepackage{caption} 
\usepackage{microtype}

\renewcommand{\paragraph}[1]{\vspace{.5em}\noindent\textbf{#1.}}



%% file: fig/teaser.tex
\begin{strip}
    \centering
    \includegraphics[width=\textwidth]{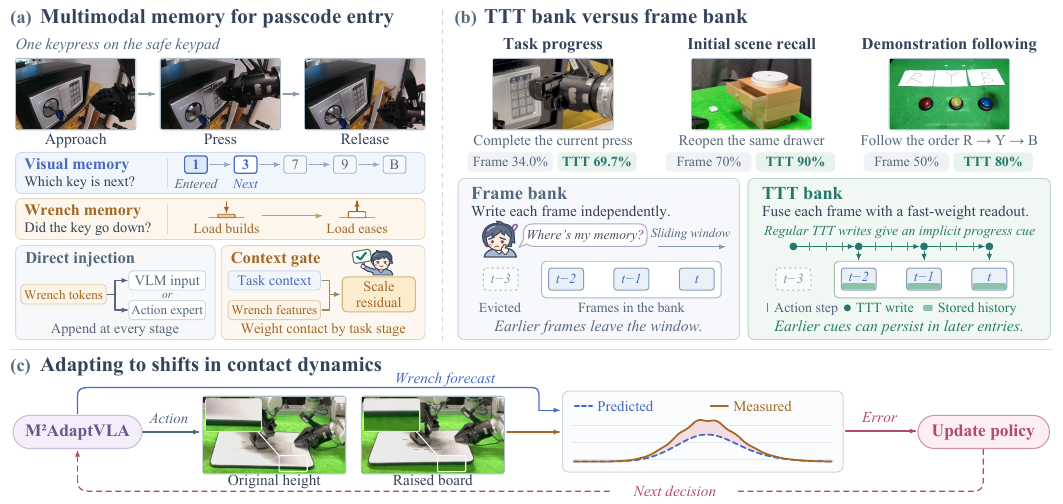}
    \captionsetup{hypcap=false}
    \captionof{figure}{\textbf{Multimodal memory and predictive test-time adaptation in M\textsuperscript{2}AdaptVLA.} (a) Visual memory cues the next key, while wrench memory signals a completed press. With direct injection, the action expert implicitly learns how much to rely on wrench memory. The context gate weights wrench history by task stage. (b) The TTT bank provides an implicit progress cue and retains history beyond the frame window, outperforming the frame bank on progress tracking, initial-scene recall, and demonstration following. (c) Raising the board changes contact dynamics. Schematic traces show how wrench forecast errors guide online adaptation to this shift.}
    \label{fig:memadaptvla_intro}
\end{strip}

%% file: sec/0_abstract.tex
    \begin{abstract}
Effective robotic manipulation calls for multimodal memory and rapid adaptation to shifts in contact dynamics.
However, many existing policies, which retain either visual or wrench history, may miss key cues from the other modality.
Moreover, even with wrench memory, fixed policies can struggle when contact conditions shift at deployment.
To this end, we introduce M\textsuperscript{2}AdaptVLA, a vision-language-action model that integrates visual and wrench memories with predictive test-time adaptation.
For visual memory, the model uses test-time training (TTT) to encode observation history in fast weights.
Each entry in the bounded TTT bank combines the corresponding frame's features with a readout from these weights, preserving cues that predate the oldest retained frame.
For wrench memory, a context gate controls how much the action expert draws on wrench history as it jointly denoises future actions and wrenches.
For predictive test-time adaptation, the auxiliary wrench prediction loss also provides online self-supervision.
Using measured wrenches as targets, we update low-rank policy adapters by backpropagating only through the final denoising step.
Real-world experiments show that the TTT bank outperforms visual memory baselines on progress tracking, scene recall, and demonstration following.
On contact-sensitive tasks, the context gate yields higher success rates than direct injection or gated cross-attention.
Predictive test-time adaptation improves robustness to shifts in contact dynamics.
On eight tasks requiring visual and wrench memory, M\textsuperscript{2}AdaptVLA achieves a mean success rate of 82.71\%, a 31.46\% absolute gain over the strongest baseline.
The project page is available at \url{https://tlf-tlf.github.io/M2AdaptVLAPage/}.
\end{abstract}

%% file: sec/1_intro.tex
\section{Introduction}
\label{sec:intro}

Recent work has incorporated memory into vision-language-action (VLA) models and demonstrated its effectiveness on robotic manipulation tasks in which the next action depends on sensory information gathered over time~\cite{memoryvla,memoact,rememvla,mem,vlamemorysurvey}. Many memory-augmented policies focus on a single sensory history, using either visual memory to retain earlier observations~\cite{tracevla,contextvla,t2mem,guhurhistory} or wrench memory to retain the forces and torques measured during contact~\cite{lee2019visiontouch,foar,fmvla}. However, a policy that relies on a single memory modality may miss key cues from the other modality, making it difficult to perform tasks that require both.

Consider a robot entering a safe's passcode (Fig.~\ref{fig:memadaptvla_intro}(a)). Visual memory helps the robot track which digits it has attempted and which key to press next. However, the gripper can look nearly identical as it approaches a key and as it lifts away after a press. Visual history alone may therefore leave the robot uncertain whether the key has gone down, causing it to move on before completing the press. Wrench memory supplies the missing cue: the rise and fall of contact force helps distinguish a completed press from an approach. This example motivates a multimodal memory-augmented VLA that integrates visual and wrench memory.

Recent multimodal VLAs such as DAM-VLA maintain fixed-size buffers for visual and force observations~\cite{damvla}. These buffers provide direct access to recent observations, but an earlier cue, such as which drawer was initially open (Fig.~\ref{fig:task_suite}(a)), can be lost once its frame leaves the window. Test-time training (TTT) extends the memory horizon by summarizing earlier observations in recurrent fast weights~\cite{ttt,robottt,t2mem,tttrm}. Compressing the entire visual history into these weights, however, can make details from a particular moment difficult to recover. For wrench memory, the task stage determines how much the policy should rely on contact feedback. Appending wrench tokens to the action expert leaves this weighting implicit~\cite{fmvla}. DAM-VLA and DeCAL make the weighting explicit by gating cross-attention with pooled force features~\cite{damvla} and a global tactile token~\cite{decal}, respectively. Because these gates depend only on contact features, their weights cannot respond directly to changes in the instruction or visual cues to task progress. Moreover, even with wrench memory, a fixed policy can struggle when contact dynamics shift at deployment. For example, raising a whiteboard changes when the eraser meets the surface and how much force it exerts along the same stroke (Fig.~\ref{fig:memadaptvla_intro}(c)).

To this end, we introduce \textbf{M\textsuperscript{2}AdaptVLA}, a VLA built on WALL-X~\cite{wallx} that combines multimodal memory with predictive test-time adaptation. For visual memory, we design a TTT bank to retain frame-specific detail while preserving cues beyond the explicit window. Each entry fuses the corresponding frame's features with a readout from the fast weights at that time (Fig.~\ref{fig:memadaptvla_intro}(b)). For wrench memory, we introduce a context gate that uses the instruction, visual context, and wrench history to determine how strongly the action expert should draw on contact information at each task stage (Fig.~\ref{fig:memadaptvla_intro}(a)). For predictive test-time adaptation, we jointly denoise future actions and wrenches to obtain a wrench forecast for each action sequence. When measurements arrive, we backpropagate the forecast error only through the final denoising step to update low-rank adapters~\cite{lora} (Fig.~\ref{fig:memadaptvla_intro}(c)).

To evaluate M\textsuperscript{2}AdaptVLA, we design eight real-robot tasks with different demands on visual and wrench memory. Our model achieves a mean task score of 82.71\%, an absolute gain of 31.46\% over the strongest baseline (51.25\%). Ablations show that the TTT bank outperforms the evaluated visual memory baselines on task progress tracking, initial-scene recall, and demonstration following. The context gate also achieves higher mean stage success than either direct wrench injection or the GCA approach. Across the two whiteboard contact shifts, predictive test-time adaptation raises mean wiping success from 85.0\% to 97.5\%, an absolute gain of 12.5\% over the fixed policy. Each update takes 159.5\,ms on average on an RTX~5090.

Our contributions are threefold.
\begin{itemize}[leftmargin=1.2em,labelsep=0.45em,itemindent=0pt,itemsep=2pt]
    \item We introduce M\textsuperscript{2}AdaptVLA, a multimodal memory-augmented VLA that combines visual memory with context-gated wrench history to jointly predict future actions and wrenches. Extensive experiments demonstrate its effectiveness.
    \item We introduce a visual memory bank that fuses each frame's features with a readout from TTT fast weights, preserving cues from beyond the bank window.
    \item We introduce a predictive test-time adaptation mechanism that updates policy adapters by backpropagating the wrench prediction loss through only the final step of joint action--wrench denoising.
\end{itemize}

%% file: sec/2_formatting.tex
\section{Related Work}
\label{sec:related_work}

\begin{figure*}[t]
    \centering
    \includegraphics[width=\textwidth]{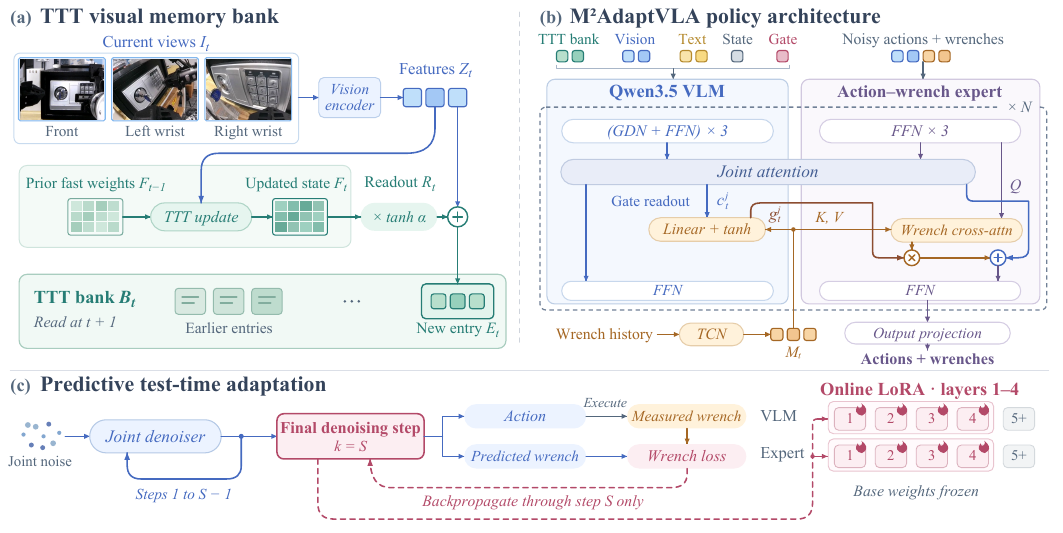}
    \caption{\textbf{M\textsuperscript{2}AdaptVLA architecture.} (a) Each bank entry combines current visual features with a gated fast-weight readout. (b) At each full-attention layer, VLM context and TCN-encoded wrench history gate the action expert's wrench residual. (c) At test time, we backpropagate the wrench loss only through the final denoising step to update LoRA adapters in layers 1--4 of the VLM and expert.}
    \label{fig:method}
\end{figure*}

\subsection{Visual Memory for Robotic Manipulation}

Visual memory helps robots track task progress~\cite{memoryvla,memoact} and recall scene cues that are no longer visible~\cite{sam2act,robomme,t2mem,liberomem,memobench,mikasa}. A common approach is to maintain a bounded memory bank~\cite{sam2act,memoryvla,memoact,damvla}. MemoryVLA stores visual features alongside instruction-conditioned cognitive tokens~\cite{memoryvla}. MemoAct extends the temporal span of a fixed-size bank by compressing older observations into long-term summaries and merging similar adjacent summaries when space runs out~\cite{memoact}. Once a frame is evicted or compressed, its details can be recovered only if the remaining entries preserve them. Sparse keyframe methods select which observations to keep~\cite{memer,eventvla,kemo,replanmem,fourdimvla}. MemER selects task-relevant frames from recent observations~\cite{memer}, while EventVLA predicts when to store incoming frames~\cite{eventvla}. A transient cue can be lost if the selector misses the frame that contains it. TTT carries history forward through recurrent updates to fast weights~\cite{ttt,robottt,t2mem,titans}, but fast weights alone provide no explicit record of a particular frame. TTT-RM trains its fast weights to capture history missing from the retrieved context and combines their readout with retrieved bank features during action generation~\cite{tttrm}. M\textsuperscript{2}AdaptVLA writes each bank entry as the sum of the current frame's features and a gated fast-weight readout, preserving frame-specific detail and cues from beyond the window.

\subsection{Wrench Memory for Robotic Manipulation}
 How much the robot needs to rely on wrench history varies across task stages~\cite{damvla,frvla,phaforce}. TA-VLA encodes torque history as a single decoder token~\cite{tavla}, and FM-VLA feeds a compressed wrench history to its action expert~\cite{fmvla}. Direct injection leaves the expert to infer when contact history matters. Fusion mechanisms make this choice explicit. DAM-VLA gates cross-attention using pooled force features alone~\cite{damvla}, so its gate has no direct access to visual cues that distinguish task stages. Other methods use predicted contact to weight force features~\cite{foar}, route multimodal features through a mixture of experts~\cite{forcevla}, or infer the contact phase from vision and language to bias attention to wrench tokens~\cite{frvla}. To weight contact history by task stage, M\textsuperscript{2}AdaptVLA combines encoded wrench history with the VLM context. Together they determine how strongly wrench history contributes to joint action--wrench denoising at each joint-attention layer.

\subsection{Online Adaptation for Robot Policies}

When contact conditions change at deployment, the same motion can produce a different wrench. Some policies adapt by inferring physical properties or latent dynamics from recent interactions~\cite{hora,scout,latentphysical,onlinesysid}. Others update parameters through self-supervision, using inverse dynamics~\cite{pad} or agreement between overlapping action predictions~\cite{salt}. For contact-rich manipulation, a predicted wrench provides a reference that the robot can check against its own measurements~\cite{forcemimic}. AdaWorldPolicy first generates actions, then predicts visual and force observations conditioned on those actions and uses the forecast errors to update low-rank adapters~\cite{adaworldpolicy}. To respond promptly to changes in contact conditions, M\textsuperscript{2}AdaptVLA predicts future actions and wrenches in a single denoising trajectory. Once the corresponding wrench measurements arrive, it limits the cost of each adapter update by backpropagating the wrench prediction loss through only the final denoising step.

%% file: sec/3_finalcopy.tex
\section{Method}
\label{sec:method}

We build M\textsuperscript{2}AdaptVLA on WALL-X~\cite{wallx} with Qwen3.5~\cite{qwen35} as its vision--language backbone.
Each of the backbone's six layer groups contains three Gated DeltaNet (GDN) blocks~\cite{gateddeltanet} followed by a full-attention block that connects the VLM to the action expert through joint attention.
At decision $t$, the VLM combines the instruction $\ell$, current views $I_t$, and proprioception $p_t$ with earlier visual observations stored in a TTT bank.
A context gate regulates how strongly recent wrench history influences the expert as it jointly predicts an action chunk $A_t$ and its corresponding wrenches $W_t^{+}$.
At deployment, we adapt the policy by backpropagating the error between predicted and measured wrenches through only the final denoising step.
Figure~\ref{fig:method} shows the full M\textsuperscript{2}AdaptVLA pipeline.

\subsection{TTT-Augmented Visual Memory Bank}
\label{sec:visual_memory}

To preserve frame-specific detail while carrying earlier cues forward, we fuse the current frame's features with a gated fast-weight readout and store the result as a bank entry (Fig.~\ref{fig:method}(a)).

Given visual features $Z_t=E_{\rm vis}(I_t)$, learned projections produce queries $q_{t,i}$, keys $k_{t,i}$, and values $v_{t,i}$ for token $i$.
Queries and keys are $\ell_2$-normalized.
The inner TTT loss measures how well the MLP with fast weights $F$ reconstructs a normalized value--key residual:
\begin{equation}
    \mathcal{L}_{t,i}^{\rm fw}(F)=\frac{1}{2}
    \big\|\operatorname{LN}(f_F(k_{t,i}))
    -\mathcal{N}(v_{t,i}-k_{t,i})\big\|_2^2,
    \label{eq:ttt_loss}
\end{equation}
where $\mathcal{N}$ standardizes the target features with a learned affine transformation and $\operatorname{LN}$ denotes layer normalization.
The current frame updates the preceding fast-weight state according to
\begin{equation}
    F_t=F_{t-1}-\sum_i\eta_{t,i}
    \nabla_F\mathcal{L}_{t,i}^{\rm fw}(F_{t-1}),
    \label{eq:ttt_update}
\end{equation}
where $\eta_{t,i}$ is the learned step size for token $i$, including mini-batch scaling. We omit the head index (see Appendix~\suppref{app:ttt_heads}{A.1} for per-head updates and readouts).

The updated fast weights are queried with a residual readout $q_{t,i}+\operatorname{LN}(f_{F_t}(q_{t,i}))$.
We merge the head outputs and project them into the visual feature space to obtain the history readout $R_t$. We then write the fused entry to the bank as follows.
\begin{equation}
\begin{aligned}
    E_t &= Z_t+\tanh(\alpha)\odot R_t,\\
    B_t &= \operatorname{FIFO}_{K}(B_{t-1}\cup\{E_t\}),
\end{aligned}
    \label{eq:bank_update}
\end{equation}
where $\alpha$ is a learned visual fusion gate and $K$ bounds the number of entries.
At time $t$, the policy reads the current features $Z_t$ and the preceding bank $B_{t-1}$. The new entry $E_t$ is available to subsequent decisions. Because later entries carry information from earlier observations, a cue can remain accessible after its original frame leaves the bank.

\subsection{Context Gate for Wrench Memory}
\label{sec:wrench_memory}

Wrist forces and torques reveal contact events that can be difficult to see. We use a causal temporal convolutional network (TCN) to encode the normalized wrench history as an ordered sequence of memory tokens $M_t$.

How strongly the policy should rely on wrench history varies across task stages. When entering a passcode, vision can guide the approach to the next key, while the contact history helps determine whether to keep pressing. We therefore use the current task context to regulate the contribution of wrench memory (see Fig.~\ref{fig:method}(b) for details). At each full-attention layer $j$, a learned read-only gate token attends to the instruction, current views, visual memory, and proprioception. We combine its readout $c_t^j$ with $M_t$ to compute a scalar gate $g_t^j$ that scales the wrench residual,
\begin{equation}
\begin{aligned}
    g_t^j &= \tanh G_j(c_t^j,M_t),\\
    \Delta H_t^j &= g_t^j\operatorname{CrossAttn}_j(H_t^j,M_t),
\end{aligned}
    \label{eq:wrench_fusion}
\end{equation}
where $H_t^j$ denotes the expert's denoising features and $G_j$ is a learned linear projection of the concatenated context and wrench-memory representation to a scalar.
Let $J_t^j$ denote the expert's native joint-attention update. The state passed to the feed-forward block is
\begin{equation}
    U_t^j = H_t^j + J_t^j + \Delta H_t^j .
    \label{eq:expert_residual}
\end{equation}

\subsection{Joint Denoising for Test-Time Adaptation}
\label{sec:joint_denoising}

We extend WALL-X's output to a joint sequence $Y_t=[A_t,W_t^{+}]$. The shared expert denoises this sequence conditioned on the multimodal VLM context $\mathcal{C}_t=(\ell,p_t,Z_t,B_{t-1})$ and the gated wrench memory $M_t$.
For Gaussian noise $\epsilon$ and flow time $\tau\in[0,1]$, we form $Y_t^\tau=(1-\tau)\epsilon+\tau Y_t$~\cite{flowmatching} and reconstruct the clean target through
\begin{equation}
    \widetilde Y_t=Y_t^\tau+(1-\tau)
    v_\theta(Y_t^\tau,\tau;\mathcal{C}_t,M_t).
    \label{eq:joint_flow}
\end{equation}

At inference, we integrate the predicted velocity from Gaussian noise to generate both sequences (see Algorithm~\suppref{alg:forward}{1} in Appendix~\suppref{app:policy_forward}{B} for the forward pass).

When contact dynamics shift at deployment, the same action can produce a different wrench. To adapt to this change, we use measured wrenches as targets for the wrench forecasts generated alongside the actions (see Fig.~\ref{fig:method}(c) for the online update). We detach the state entering the final denoising step and replay that step to obtain $\widehat W_t^{+}(\phi)$ under the LoRA adapters $\phi$~\cite{lora}. We update the adapters with
\begin{equation}
    \mathcal{L}_{\rm adapt}(\phi)=
    \|\widehat W_t^{+}(\phi)-W_t^{\rm meas}\|_{\Omega_t}^{2},
    \label{eq:adaptation_loss}
\end{equation}
where $\Omega_t$ selects valid wrench measurements aligned with the executed actions (see Appendix~\suppref{app:wrench_feedback}{C} for the selection criteria). We average the squared error over the selected coordinates using the training normalization and update only the adapters in the first four layers of the VLM and expert.

\section{Experiments}
\label{sec:experiments}

To comprehensively evaluate the performance of M\textsuperscript{2}AdaptVLA, we conducted a series of real-robot experiments around the following four questions:

\textbf{(Q1)} How does M\textsuperscript{2}AdaptVLA compare with state-of-the-art VLA baselines on tasks that rely on visual and wrench memory?

\textbf{(Q2)} What advantages does the TTT bank offer over existing visual memory methods?

\textbf{(Q3)} Does the context gate outperform existing methods for injecting wrench history into the policy?

\textbf{(Q4)} Can a policy with predictive test-time adaptation better handle shifts in contact dynamics at deployment than a fixed policy?

\subsection{Experimental Setup}
\label{sec:experimental_setup}

\begin{figure}[H]
    \centering
    \includegraphics[width=\columnwidth]{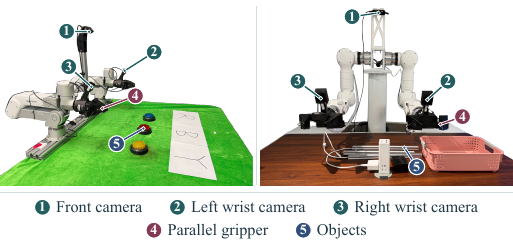}
    \caption{\textbf{Robot platforms.} The TeleTwin Desktop (left) and Tianji (right) each have one front camera and two wrist cameras.}
    \label{fig:robot_platforms}
\end{figure}

\paragraph{Robot platforms}
We use two bimanual platforms (Fig.~\ref{fig:robot_platforms}). The TeleTwin Desktop from X Square Robot has two six-axis arms, and the Tianji has two seven-axis arms. Both provide images from a fixed front camera and a camera on each wrist, along with proprioception and six-axis wrist wrench measurements.

\begin{figure*}[!t]
    \centering
    \includegraphics[width=\textwidth]{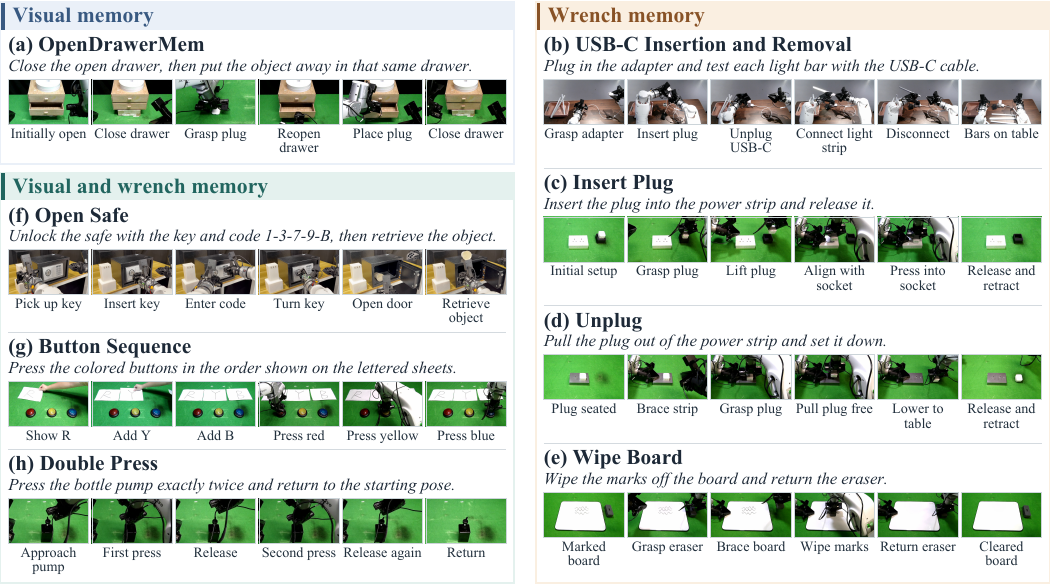}
    \caption{\textbf{Eight real-robot tasks grouped by memory capability.} Panels (a), (b--e), and (f--h) test visual memory, wrench memory, and both, respectively. }
    \label{fig:task_suite}
\end{figure*}

\paragraph{Training and inference}
We train M\textsuperscript{2}AdaptVLA on eight NVIDIA A800 GPUs. At deployment, we run inference and online adaptation on a single NVIDIA RTX~5090.

\paragraph{Tasks}
We designed eight real-robot tasks that rely on visual memory, wrench memory, or both, with 5,953 demonstrations in total (see Table~\ref{tab:real_tasks} for per-task counts). Figure~\ref{fig:task_suite} shows the instruction and a sequence of demonstration frames for each task. We group the tasks by the memory capability they test: visual memory, wrench memory, or both.

In OpenDrawerMem, the robot closes the initially open drawer, picks up an object, and uses visual memory to reopen the same drawer. Four tasks probe wrench memory. In USB-C Insertion and Removal, Insert Plug, and Unplug, changes in wrist force and torque indicate when a connector seats or comes free. In Wipe Board, wrench history helps maintain contact throughout each stroke. The remaining three tasks require both modalities. The robot uses visual memory to track passcode progress in Open Safe, recall the demonstrated order in Button Sequence, and count pump strokes in Double Press. In all three, wrench feedback indicates when each press is complete.

\paragraph{Baselines}
\textbf{(Q1)} We use $\pi_{0.5}$~\cite{pi05}, a widely used generalist VLA, as our reference policy. We also compare against two VLAs with visual memory, $\pi$-Mem (the visual memory variant of MEM~\cite{mem}) and MemoryVLA~\cite{memoryvla}, and two force-aware VLAs, ForceVLA~\cite{forcevla} and TA-VLA~\cite{tavla}.

\begin{figure}[!htbp]
    \centering
    \includegraphics[width=\columnwidth]{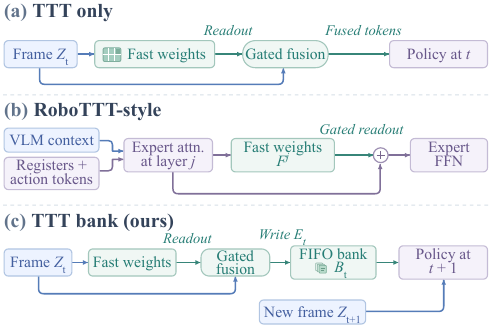}
    \caption{\textbf{Where TTT memory enters the policy.} (a) The fast-weight readout enters the current visual input. (b) Layer-local fast weights feed a gated residual back into the action expert before its feed-forward block. (c) A bank entry written at decision $t$ first becomes available at $t+1$, when it can be read alongside fresh visual features.}
    \label{fig:visual_memory_variants}
\end{figure}

\textbf{(Q2)} We compare visual memory designs on the same WALL-X backbone~\cite{wallx}. \emph{Frame Bank} stores independently encoded frames in a first-in, first-out buffer, and \emph{GDN} carries the native Gated DeltaNet state~\cite{gateddeltanet} of Qwen3.5~\cite{qwen35} across decisions. Figure~\ref{fig:visual_memory_variants} compares the three designs that use TTT. The \emph{TTT only} variant fuses the fast-weight readout with current visual features and uses the result at the same decision. The \emph{RoboTTT-style} implementation~\cite{robottt} places learned register tokens in the action expert and applies TTT to expert features after attention. Both variants carry history in fast weights. Our \emph{TTT bank} also stores each fused visual representation in a bounded bank. At the next decision, the policy reads these entries alongside the new camera views. All variants are trained on chronologically ordered decisions within each demonstration (Appendix~\suppref{app:temporal_training}{A}).

\textbf{(Q3)} To evaluate the context gate, we keep the TTT bank fixed and vary only how wrench history enters the policy. Two variants append wrench tokens directly to the VLM input or to the action expert, the latter following TA-VLA~\cite{tavla}. A third uses gated cross-attention (GCA) with a gate computed from pooled wrench features, following DAM-VLA~\cite{damvla}.

\begin{figure}[!t]
    \centering
    \includegraphics[width=\columnwidth]{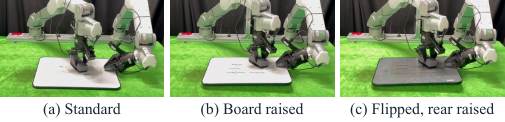}
    \caption{\textbf{Whiteboard contact conditions.} The board is (a) flat at its original height, (b) raised, or (c) flipped with its rear edge raised.}
    \label{fig:whiteboard_shifts}
\end{figure}

\textbf{(Q4)} We test the same wiping policy with and without predictive test-time adaptation under two contact conditions (see Fig.~\ref{fig:whiteboard_shifts}): we either raise the entire board or flip it and raise its rear edge.

\begin{table*}[!t]
\centering
\caption{\textbf{Success rates (\%) on eight real-robot tasks.} We report each method's per-task success rates and their mean. Bold marks the highest value in each row, and underlining marks the next highest distinct value.}
\label{tab:real_tasks}
\small
\setlength{\tabcolsep}{2pt}
\begin{tabular*}{\textwidth}{@{\extracolsep{\fill}}lrrrrrrr@{}}
\toprule
Task & Demos & $\pi_{0.5}$~\cite{pi05} & $\pi$-Mem~\cite{mem} & MemoryVLA~\cite{memoryvla} & ForceVLA~\cite{forcevla} & TA-VLA~\cite{tavla} & M\textsuperscript{2}AdaptVLA\\
\midrule
Open Safe       & 2,011 & 0 & 0 & \underline{26.7} & 3.3 & 0 & \textbf{66.7}\\
USB-C Insertion and Removal & 454 & 25 & 30 & 15 & \underline{50} & 40 & \textbf{65}\\
Insert Plug     & 500 & 55 & 70 & 15 & 85 & \underline{90} & \textbf{100}\\
Unplug          & 500 & \textbf{100} & 90 & \underline{95} & \textbf{100} & \textbf{100} & \textbf{100}\\
Wipe Board      & 500 & 30 & 45 & 0 & \underline{50} & 45 & \textbf{100}\\
OpenDrawerMem   & 627 & 40 & \underline{75} & 50 & 50 & 25 & \textbf{90}\\
Button Sequence & 829 & 0 & \underline{55} & \underline{55} & 0 & 0 & \textbf{80}\\
Double Press    & 532 & 20 & \underline{45} & 10 & 20 & 10 & \textbf{60}\\
\midrule
Eight-task mean & --- & 33.75 & \underline{51.25} & 33.33 & 44.79 & 38.75 & \textbf{82.71}\\
\bottomrule
\end{tabular*}
\end{table*}

\paragraph{Evaluation protocol}
In the Open Safe ablations, we score each of the nine stages separately so that an early failure does not mask the policy's ability to complete later stages. To reduce sampling variability, we estimate each stage's success rate from 60 trials per variant. For whole-task evaluation, we run 30 trials per policy on Open Safe and 20 on each other task, counting a trial as successful only if the robot completes all required stages.

\begin{table*}[!t]
\centering
\caption{\textbf{Open Safe ablations by stage (\%).} Each stage is evaluated over 60 trials per variant. Press SR and Mean SR average the five keypad stages and all nine stages, respectively. Ranks are within groups.}
\label{tab:safe_stages}
\begingroup\small
\newcommand{\oursrow}{%
  \rlap{\smash{\color[gray]{0.93}%
    \rule[-\dp\csname @arstrutbox\endcsname]{\textwidth}%
      {\dimexpr\ht\csname @arstrutbox\endcsname+\dp\csname @arstrutbox\endcsname\relax}}}}
\setlength{\tabcolsep}{2.5pt}
\begin{tabular*}{\textwidth}{@{\extracolsep{\fill}}lrrrrrrrrrrr@{}}
\toprule
& \multicolumn{2}{c}{Key} & \multicolumn{5}{c}{Keypad} &&&&\\
\cmidrule(lr){2-3}\cmidrule(lr){4-8}
Variant & Grasp & Insert & 1 & 3 & 7 & 9 & B & Turn & Retrieve & Press SR & Mean SR\\
\midrule
\multicolumn{12}{@{}l}{\textit{Visual memory}}\\
Frame Bank & \underline{96.7} & 21.7 & \underline{31.7} & 35.0 & \underline{35.0} & \underline{33.3} & 35.0 & 90.0 & \underline{96.7} & 34.0 & 52.8\\
RoboTTT-style~\cite{robottt} & \underline{96.7} & \underline{33.3} & 20.0 & 23.3 & 10.0 & 3.3 & 6.7 & 83.3 & 93.3 & 12.7 & 41.1\\
GDN~\cite{gateddeltanet} & \textbf{100.0} & 6.7 & 30.0 & \underline{70.0} & 20.0 & 18.3 & \underline{36.7} & \textbf{100.0} & \textbf{100.0} & \underline{35.0} & 53.5\\
TTT only & \textbf{100.0} & \textbf{70.0} & \underline{31.7} & 43.3 & 20.0 & 23.3 & 31.7 & \textbf{100.0} & \textbf{100.0} & 30.0 & \underline{57.8}\\
\oursrow TTT bank (ours) & 91.7 & \textbf{70.0} & \textbf{73.3} & \textbf{73.3} & \textbf{71.7} & \textbf{61.7} & \textbf{68.3} & \underline{91.7} & 88.3 & \textbf{69.7} & \textbf{76.7}\\
\midrule
\multicolumn{12}{@{}l}{\textit{Wrench history injection with the TTT bank}}\\
VLM input & 91.7 & \underline{53.3} & 8.3 & 15.0 & 3.3 & 0.0 & 0.0 & 18.3 & \underline{93.3} & 5.3 & 31.5\\
Action expert input & \textbf{100.0} & 51.7 & \underline{68.3} & \underline{61.7} & \textbf{71.7} & \textbf{63.3} & \underline{38.3} & 85.0 & 85.0 & \underline{60.7} & \underline{69.4}\\
GCA~\cite{damvla} & \underline{98.3} & 38.3 & 21.7 & 41.7 & \underline{18.3} & \underline{23.3} & 21.7 & \underline{88.3} & 85.0 & 25.3 & 48.5\\
\oursrow Context gate (ours) & 93.3 & \textbf{75.0} & \textbf{73.3} & \textbf{76.7} & \textbf{71.7} & \textbf{63.3} & \textbf{70.0} & \textbf{98.3} & \textbf{95.0} & \textbf{71.0} & \textbf{79.6}\\
\midrule
\multicolumn{12}{@{}l}{\textit{Future wrench prediction with the TTT bank and context gate}}\\
Without prediction & \underline{93.3} & \underline{75.0} & \underline{73.3} & \textbf{76.7} & \underline{71.7} & \underline{63.3} & \underline{70.0} & \textbf{98.3} & \underline{95.0} & \underline{71.0} & \underline{79.6}\\
\oursrow With prediction (ours) & \textbf{98.3} & \textbf{76.7} & \textbf{75.0} & \textbf{76.7} & \textbf{80.0} & \textbf{70.0} & \textbf{76.7} & \textbf{98.3} & \textbf{98.3} & \textbf{75.7} & \textbf{83.3}\\
\bottomrule
\end{tabular*}\endgroup
\end{table*}

\subsection{Experimental Results}
\label{sec:baseline_comparison}

\paragraph{M\textsuperscript{2}AdaptVLA outperforms state-of-the-art VLA baselines~(Q1)}
Effective manipulation across these tasks calls for both visual and wrench memory. A single memory modality can capture the evidence needed for one task yet miss a decisive cue in another. As shown in Table~\ref{tab:real_tasks}, the relative strengths of the baselines reverse across two tasks. On Button Sequence, the robot must recall the demonstrated order. Both visual memory baselines reach 55\% success, while both force-aware baselines score 0\%. On USB-C Insertion and Removal, wrist feedback indicates when a connector seats or comes free. Here the force-aware baselines reach 40--50\%, compared with 15--30\% for the visual memory baselines. M\textsuperscript{2}AdaptVLA performs best on both tasks, achieving 80\% and 65\%, respectively.

Open Safe requires both histories: visual memory tracks the next key, while wrench history indicates whether a press is complete. M\textsuperscript{2}AdaptVLA achieves 66.7\% success, compared with 26.7\% for MemoryVLA, the strongest baseline on this task. Across eight tasks, our policy ranks first on seven and ties for first on Unplug, achieving 82.71\% mean success, an absolute gain of 31.46\% over $\pi$-Mem (51.25\%).

\subsection{Ablation Studies}
\label{sec:ablations}

\paragraph{The TTT bank improves progress tracking, initial-scene recall, and demonstration following~(Q2)}
\label{sec:memory_experiments}
The TTT bank improves progress tracking during passcode entry. On Open Safe, it completes 69.7\% of individual keypresses, compared with 35.0\% for GDN, 34.0\% for Frame Bank, and 30.0\% for TTT only (Table~\ref{tab:safe_stages}). For example, in failed Frame Bank trials, the robot often moved to the next key before pressing the current one. Because the bank stores each frame independently, the policy must infer progress from views that look nearly identical just before a press and just after release.

By comparison, TTT only and RoboTTT-style advanced prematurely less often, but their presses frequently missed the small target key. Both variants update their fast weights after a fixed number of action steps (Fig.~\ref{fig:memadaptvla_intro}(b)). This implicit progress cue may explain why they were less likely than Frame Bank to move on before attempting a press. However, both variants compress visual history into fast weights and integrate the readout with features from the current frame, either in the visual input or in the action expert. Without a separate bank of historical features, the policy may have difficulty recovering the fine spatial cues needed to align the gripper with a small key. M\textsuperscript{2}AdaptVLA retains the progress cue while supplying the current views separately from the historical bank, allowing the policy to align the gripper directly from those views.

\begin{table}[!htbp]
\centering
\caption{\textbf{Visual memory for drawer recall and button-sequence following.} Whole-task success (\%) over 20 trials per condition.}
\label{tab:visual_tasks}
\small
\setlength{\tabcolsep}{2.5pt}
\newcommand{\ourrowshade}{\rlap{\smash{\color[gray]{0.93}\rule[-\dp\csname @arstrutbox\endcsname]{\columnwidth}{\dimexpr\ht\csname @arstrutbox\endcsname+\dp\csname @arstrutbox\endcsname\relax}}}}
\begin{tabular*}{\columnwidth}{@{\extracolsep{\fill}}lrr@{}}
\toprule
Visual memory & OpenDrawerMem & Button Sequence\\
\midrule
Frame Bank & 70 & 50\\
TTT only & \underline{75} & 55\\
RoboTTT-style~\cite{robottt} & 55 & \underline{60}\\
GDN~\cite{gateddeltanet} & 55 & 55\\
\ourrowshade TTT bank~(ours) & \textbf{90} & \textbf{80}\\
\bottomrule
\end{tabular*}
\end{table}

Beyond tracking an ongoing action, the TTT bank also helps the robot recall the initial scene. On OpenDrawerMem, the robot must reopen the drawer that was initially open. As shown in Table~\ref{tab:visual_tasks}, the TTT bank achieves 90\% whole-task success, compared with 75\% for TTT only, 70\% for Frame Bank, and 55\% for RoboTTT-style. The bank can preserve the initial drawer configuration because each entry combines the visual features of a particular frame with a readout of the history up to that point (Eq.~\ref{eq:bank_update}). Later entries can thus retain cues about which drawer was open even after the initial view leaves the bank.

The advantage extends from recalling an initial scene to following a demonstrated sequence. On Button Sequence, the robot must remember the order in which the demonstrator presented the button cues. With the TTT bank, it reproduces that order in 80\% of trials, compared with 60\% for RoboTTT-style, the strongest alternative on this task.

\paragraph{The context gate outperforms direct wrench injection and GCA~(Q3)}
\label{sec:wrench_experiments}
 As shown in Table~\ref{tab:safe_stages}, with the same TTT bank, the context gate achieves 79.6\% mean stage success on Open Safe, compared with 69.4\% for direct injection into the action expert, 31.5\% for direct injection into the VLM, and 48.5\% for GCA. Direct injection makes wrench tokens available at every stage but leaves their relevance to be inferred through attention. Explicitly weighting their contribution improves success over direct expert input from 51.7\% to 75.0\% for key insertion and from 38.3\% to 70.0\% for the final keypad press.

\begin{figure}[!htbp]
    \centering
    \includegraphics[width=\columnwidth]{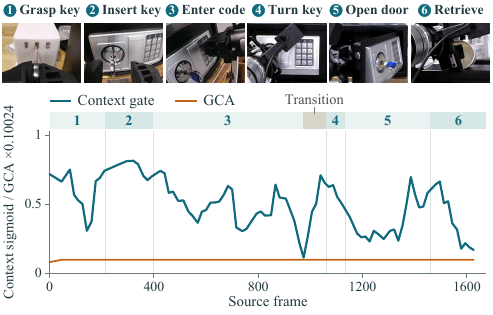}
    \caption{Context gate and GCA gate values across stages of the same Open Safe episode.}
    \label{fig:gate_stage_replay}
\end{figure}

Figure~\ref{fig:gate_stage_replay} compares the two gates on the same Open Safe episode. The context gate gives wrench history greater weight as the robot inserts the key, when resistance can reveal misalignment, and less weight as it swings the unlocked door open. This variation suggests that the policy learns to use contact feedback selectively as the task progresses. However, because GCA computes its gate from pooled wrench features alone, it cannot adjust the weight directly in response to visual cues about task progress. Its gate weight remained nearly constant across stages in this episode.

In addition, using future wrench prediction as auxiliary supervision raises keypad press success from 71.0\% to 75.7\% and mean stage success from 79.6\% to 83.3\%. These gains show that training the policy to predict actions and wrenches jointly can improve task performance.

\paragraph{Predictive test-time adaptation improves robustness to shifts in contact dynamics~(Q4)}
\label{sec:online_adaptation_eval}
Raising or flipping the contact surface changes the load on the eraser during wiping. Both policies observe wrist wrenches, but the adaptive policy also uses the difference between predicted and measured wrenches to update its adapters. As shown in Table~\ref{tab:wipe_adaptation}, this raises whole-task success from 90\% to 100\% on the raised board and from 80\% to 95\% on the flipped board with its rear edge raised.

\begin{table}[!htbp]
\centering
\caption{\textbf{Predictive test-time adaptation on Wipe Board.} Success rates are based on 20 trials per policy and condition. Parentheses give the relative change ($\Delta$) from the fixed policy. LoRA variants were timed sequentially on one RTX~5090.}
\label{tab:wipe_adaptation}
\small
\setlength{\tabcolsep}{2pt}
\begin{tabular*}{\columnwidth}{@{\extracolsep{\fill}}lrrr@{}}
\toprule
 & Fixed & LoRA 1--4 & LoRA 1--24 \\
\midrule
Board raised (\%) & \underline{90} & \textbf{100}\,{\scriptsize $(+11.1\%)$} & 80\,{\scriptsize $(-11.1\%)$} \\
Flipped, rear raised (\%) & \underline{80} & \textbf{95}\,{\scriptsize $(+18.8\%)$} & \textbf{95}\,{\scriptsize $(+18.8\%)$} \\
Mean success (\%) & 85.0 & \textbf{97.5}\,{\scriptsize $(+14.7\%)$} & \underline{87.5}\,{\scriptsize $(+2.9\%)$} \\
\midrule
Trainable parameters (M) & 0 & 3.18 & 19.08 \\
Time per update (ms) & n/a & 159.5 & 267.7 \\
\bottomrule
\end{tabular*}
\end{table}

Adapting more layers did not improve robustness. LoRA in the first four layers of both the VLM and action expert achieved 97.5\% mean success, compared with 87.5\% for all 24 layers and 85.0\% for the fixed policy. The two adaptive variants tied on the flipped board, but only the four-layer variant improved on the fixed policy when the entire board was raised. Four-layer updates take 159.5\,ms on an RTX~5090, compared with 267.7\,ms for all layers.

\section{Conclusion}
\label{sec:conclusion}
In this work, we introduced M\textsuperscript{2}AdaptVLA, a vision-language-action policy that integrates visual and wrench memories with predictive test-time adaptation.
In real-robot experiments, its TTT bank outperformed visual memory baselines in tracking task progress, recalling the initial scene, and following demonstrated sequences.
The context gate adjusted the influence of wrench history across task stages.
Predictive test-time adaptation improved robustness to changes in contact conditions with low update latency.
Across eight real-robot tasks, M\textsuperscript{2}AdaptVLA achieved 82.71\% mean success, an absolute gain of 31.46\% over the strongest baseline.
A limitation of our approach is that maintaining both visual and wrench memories increases overall inference latency.
In future work, we will explore a two-timescale architecture in which a slow module updates multimodal memory and a fast controller uses the latest memory and current observations to select actions in real time.

\section*{LLM Usage Statement}
We used OpenAI's ChatGPT and Codex to assist with drafting and restructuring the manuscript, editing its language, refining mathematical arguments and exposition, discussing the method, checking literature sources and references, reviewing successive drafts, and writing visualization scripts. The authors take responsibility for the paper's claims, citations, figures, and results.

%% file: sec/X_suppl_from_original.tex
\clearpage
\appendix
\maketitlesupplementary

\section{Temporal Training Details}
\label{app:temporal_training}

\paragraph{Sampling the Open Safe training context}
For the Open Safe ablations, we train all five visual memory variants on chronologically ordered decisions. After filtering stationary frames, we sample a contiguous window of 1,000 frames per demonstration each epoch, using the full filtered sequence for shorter demonstrations. We select the first frame and then every 16th frame, yielding at most 63 decisions per window. Frame Bank retains independently encoded frames. The TTT bank, TTT only, GDN, and RoboTTT-style variants carry their respective memory states across decisions.

\paragraph{Truncating gradients while retaining history}
Each GPU processes up to 12 successive decisions from one window per segment, giving at most 96 decision positions across eight GPUs. A full 63-decision window therefore spans six segments. Only valid supervised positions contribute to the loss. We initialize the fast weights and FIFO bank at the start of each sampled window and carry their values across segments with gradients detached. We replay the preceding segment's final bank write from its detached input so that the next segment can supervise that write. Gradients accumulate across segments; the outer optimizer takes one step when the windows finish. Thus, 12 bounds the gradient horizon without limiting the history available for prediction.

\subsection{Head-wise TTT update and readout}
\label{app:ttt_heads}

At decision $t$, we group the $m$ visual tokens from all synchronized views into one batch. The TTT module has eight heads. Each head $h$ maintains the weights and biases $F^h$ of its own two-layer MLP $f_{F^h}$ with a GELU activation. Projecting a visual token $z_{t,i}$ gives $q_{t,i}^{h}$, $k_{t,i}^{h}$, and $v_{t,i}^{h}$. We normalize each query and key to unit length within its head.

For head $h$, the loss $\mathcal{L}_{t,i}^{h}$ is Eq.~\eqref{eq:ttt_loss} evaluated with that head's key, value, and normalization parameters. All tokens in the batch use the same incoming state $F_{t-1}^{h}$. With head width $d_h$, base step size $\eta_0$, and learned parameters $u_h,b_h$, the update is
\begin{align}
  \eta_{t,i}^{h}
    &= \frac{\eta_0}{m d_h}\,\sigma(u_h^\top z_{t,i}+b_h), \\
  F_t^{h}
    &= F_{t-1}^{h}
       -\sum_{i=1}^{m}\eta_{t,i}^{h}
        \nabla_{F^h}\mathcal{L}_{t,i}^{h}(F_{t-1}^{h}).
\end{align}
Here $\sigma$ is the sigmoid. After updating, each head reads its new state with the query from the same token. We concatenate the eight readouts, apply layer normalization, and project back to the visual feature space,
\begin{align}
  r_{t,i}^{h} &= q_{t,i}^{h}
       +\operatorname{LN}_{h}\!\left(f_{F_t^{h}}(q_{t,i}^{h})\right), \\
  R_{t,i} &= P_o\!\left(\operatorname{LN}_{\rm out}
        \big([r_{t,i}^{1},\ldots,r_{t,i}^{8}]\big)\right).
\end{align}
The fused token $z_{t,i}+\tanh(\alpha)\odot R_{t,i}$ enters the bank via Eq.~\eqref{eq:bank_update}. The head states $F_t^h$ carry forward to the next decision.

\subsection{Offline objective and hyperparameters}

The Open Safe policy uses a Qwen3.5 backbone with 2.23 billion parameters, including its vision encoder and tied embeddings. The action--wrench expert has 351.5 million parameters, including its joint input and output projections and its time projection. Both counts exclude the added memory and fusion modules. Table~\ref{tab:supp_offline} lists the training settings.

The visual inner loss in Eq.~\eqref{eq:ttt_loss} updates the fast state. The outer imitation objective supervises the joint denoiser. Its output is a velocity, but we compute the action and wrench losses on the reconstructed clean endpoint in Eq.~\eqref{eq:joint_flow}. For a normalized target $Y_t=[A_t,W_t^+]$ and noise $\epsilon$, the endpoint error equals $(1-\tau)[v_\theta-(Y_t-\epsilon)]$. The implemented squared error therefore weights the corresponding velocity error by $(1-\tau)^2$. We compute a separate mean over valid coordinates for each modality and combine the two losses as $\mathcal{L}_A+0.1\,\mathcal{L}_W$; padded action coordinates and unavailable wrench targets do not contribute.

\begin{table}[H]
\centering
\caption{Offline training settings for the Open Safe policy with joint action--wrench prediction.}
\label{tab:supp_offline}
\small
\setlength{\tabcolsep}{3pt}
\begin{tabular*}{\columnwidth}{@{\extracolsep{\fill}}ll@{}}
\toprule
Setting & Value \\
\midrule
Training window / decision stride & 1,000 / 16 frames \\
TBPTT segment / FIFO capacity & 12 decisions / 5 entries \\
Visual tokens per bank entry & $3$ views $\times$ $70$ tokens \\
TTT heads / base inner step $\eta_0$ & 8 / 0.1 \\
Wrench history & 10 samples, 12 axes each \\
Joint prediction horizon & 32 action--wrench pairs \\
Action / wrench channels & $20+6$ padding / 12 \\
Optimizer / gradient norm clip & DMuon / 1.0 \\
VLM and expert learning rate & $10^{-4}$ for both \\
LR schedule & 500-step warmup; constant \\
\bottomrule
\end{tabular*}
\end{table}

\clearpage
\subsection{From a visual observation to a bank entry}
\label{app:explicit_write}

\paragraph{One joint write for the three views}
Let $Z_t\in\mathbb{R}^{m\times D}$ be the visual features at decision $t$. In the joint policy, $D=2048$ and $m=210$: each of the front, left-wrist, and right-wrist cameras contributes 70 tokens. The tokens from all three views update the same incoming state. Updating the fast weights once per camera would make the write depend on the order of the views; the model instead sums the token contributions before changing the weights.

\paragraph{Reconstruction target}
We give the write for one of the eight heads, suppressing the indices $h$ and $t$. Tokens are row vectors of width $d=256$. The key $k_i$ is unit-normalized; the value $v_i$ is projected without this normalization. Set $u_i=v_i-k_i$. The target $y_i$ and the normalized prediction $N(o_i)$ use the same learned channel scale $\gamma$ and bias $\beta$:
\begin{align}
y_i &= \gamma\odot\frac{u_i-\bar u_i}{s(u_i)+\varepsilon}+\beta,\nonumber\\
N(o_i)&=\gamma\odot\frac{o_i-\bar o_i}
 {\sqrt{\operatorname{var}(o_i)+\varepsilon}}+\beta .
\label{eq:supp_ttt_norm}
\end{align}
Here $\bar u_i$ and $\bar o_i$ are channel means, $s$ is the sample standard deviation, $\operatorname{var}$ is the population variance, and $\varepsilon=10^{-8}$. The distinction matters when reproducing the update from a library normalization routine. The inner objective is $\tfrac12\|N(o_i)-y_i\|_2^2$.

\paragraph{Updating the two fast layers}
Both fast matrices are $d\times d$ and each has a bias of length $d$. For the incoming fast parameters, define
\begin{align}
 a_i &= k_i W_1+b_1, & h_i &= \phi(a_i),\nonumber\\
 o_i &= h_i W_2+b_2, & g_i &= \nabla_{o_i}\tfrac12\|N(o_i)-y_i\|_2^2 .
 \label{eq:supp_fast_mlp}
\end{align}
Here $\phi$ is the tanh approximation to GELU. Write $e_i=(g_iW_2^\top)\odot\phi'(a_i)$. With the token step sizes $\eta_i$ from Appendix~\ref{app:ttt_heads}, the update in Eq.~\eqref{eq:ttt_update} becomes
\begin{align}
 W_1' &= W_1-\sum_i\eta_i k_i^\top e_i,
 & b_1' &= b_1-\sum_i\eta_i e_i,\nonumber\\
 W_2' &= W_2-\sum_i\eta_i h_i^\top g_i,
 & b_2' &= b_2-\sum_i\eta_i g_i .
 \label{eq:supp_fast_gradient}
\end{align}
All right-hand sides use the same pre-write weights; in particular, $e_i$ uses $W_2$, not $W_2'$. The unit-normalized query reads from the updated MLP. Thus the readout stored at $t$ has already incorporated the evidence from $t$, but it does not enter the policy's bank input until the next decision.

\paragraph{How future actions supervise a stored entry}
The bank writes a frame feature and its fast-weight readout as a single vector. For token $i$ and visual channel $c$,
\begin{equation}
 E_{t,ic}=Z_{t,ic}+a_cR_{t,ic},
 \qquad a_c=\tanh(\alpha_c).
 \label{eq:supp_channel_write}
\end{equation}
The learned $a_c$ is shared across views, patch locations and decision times. It is distinct from the scalar, task-conditioned wrench gate in Eq.~\eqref{eq:wrench_fusion}. Let $G_{ic}$ be the derivative of a later supervised loss with respect to this stored entry on a connected training path. The local derivatives for this write are
\begin{align}
 \frac{\partial\mathcal L}{\partial R_{t,ic}}\bigg|_{E_t}
    &=a_c G_{ic},\nonumber\\
 \frac{\partial\mathcal L}{\partial\alpha_c}\bigg|_{E_t}
    &=(1-a_c^2)\sum_i G_{ic} R_{t,ic}.
 \label{eq:supp_write_credit}
\end{align}
These expressions describe how the action--wrench objective can train the contribution of a historical readout. The action at $t$ does not supervise its own bank write through this path: its visual prefix contains $Z_t$ and the earlier entries in $B_{t-1}$. The gradient above becomes available when a later prediction reads $E_t$.

\paragraph{What remains after eviction}
With five entries the FIFO holds at most 1,050 historical visual tokens, in addition to 210 current tokens outside the bank. For example, deleting $E_s$ removes direct access to its patch features. It does not reset the fast weights used to write $E_{s+1}$. A later entry can therefore retain evidence from $s$, without guaranteeing that the original frame can be reconstructed. This distinction is particularly relevant in OpenDrawerMem: the policy has no permanently reserved copy of the initially open drawer.

\paragraph{Memory capacity at inference}
Two quantities bound the visual memory cost. For $H_v$ heads of width $d$, the trajectory-specific fast state contains $H_v(2d^2+2d)$ scalars, counting both biases. It is updated in place of the preceding state and does not grow with the number of decisions. The explicit bank, by contrast, contributes one set of $m$ visual embeddings to the VLM prefix for each retained decision. At decision $t$ after $t$ prior writes, its visual-token count is $m\min(t,K)$; camera labels and temporal markers are additional. Increasing the rollout length beyond $K$ therefore does not increase the number of historical patch tokens in the prefix.

\begin{table}[H]
\centering\small
\caption{Visual memory capacity in the deployed policy. Counts describe one active rollout and exclude frozen model weights.}
\setlength{\tabcolsep}{3pt}
\begin{tabular*}{\columnwidth}{@{\extracolsep{\fill}}lr@{}}
\toprule
Quantity & Size \\
\midrule
Fast-state scalars ($H_v=8$, $d=256$) & 1,052,672 \\
Historical patch tokens ($K=5$, $m=210$) & up to 1,050 \\
Current patch tokens (all views) & 210 \\
\bottomrule
\end{tabular*}
\end{table}

\clearpage
\subsection{Learning across a segment boundary}
\label{app:credit_assignment}

\paragraph{Carrying memory without carrying its graph}
An absolute fast weight contains two quantities with different lifetimes: a learned initialization $F_\star$, shared across demonstrations, and an increment $D_t$ that belongs to this trajectory. The implementation stores the increment,
\begin{equation}
 F_t=F_\star+D_t,\qquad D_{\rm init}=0.
 \label{eq:supp_delta_state}
\end{equation}
This applies to both matrices and both biases of each head. At a truncation boundary, the next segment starts from $F_\star+\operatorname{sg}(D_t)$, where $\operatorname{sg}$ denotes stop-gradient. This preserves the numerical state accumulated earlier. It also leaves a direct gradient path to $F_\star$ without differentiating through the earlier sequence of increments.

\paragraph{Why replay the final write?}
The final observation of a segment writes a new entry, but the policy first reads that entry in the \emph{next} segment. Simply detaching the entire bank would disconnect the first such read from the operation that produced the entry. Let $b$ denote the last valid write of the preceding segment, let $S=(D,B)$, and let $\mathcal U_\theta$ denote a complete visual write (fast update, readout, fusion and FIFO insertion). At the start of the next segment we reconstruct
\begin{align}
 \bar S_{b-1} &=
   \big(\operatorname{sg}(D_{b-1}),\operatorname{sg}(B_{b-1})\big),\nonumber\\
 \widehat S_b &= \mathcal U_\theta\!\left(\bar S_{b-1},
              E_{\rm vis}(I_b^{\rm aug})\right).
 \label{eq:supp_boundary_replay}
\end{align}
We save the pre-write state and the actual augmented camera inputs, not just the resulting feature vector. The new segment re-encodes the same pixels with the same model parameters. Its starting state is the replayed $\widehat S_b$, which contains both the new fast-state increment and the updated FIFO. Thus its loss can differentiate through the final write and image encoding. Entries already present in $\bar S_{b-1}$ remain detached. Replay neither inserts $E_b$ twice nor introduces a second action target at $b$.

\par\smallskip\noindent\begin{minipage}{\columnwidth}
\hrule\smallskip
\textbf{Training procedure.} One outer update for parallel demonstration windows.
\smallskip\hrule\smallskip
\small
\expandafter\def\csname theHALG@line\endcsname{train.\arabic{ALG@line}}
\begin{algorithmic}[1]
\Require Chronological windows, segment size 12; let $n_j$ be the number of supervised decisions in segment $j$
\State Initialize episode states; zero outer gradients
\For{each segment $j$ in temporal order}
    \If{a boundary write was saved}
        \State Re-encode its images; restore $\widehat S_b$ by Eq.~\eqref{eq:supp_boundary_replay}
    \EndIf
    \For{each valid decision $t$ in segment $j$}
        \State Read the previous FIFO and the current observation
        \State Evaluate the joint objective if $t$ is supervised
        \State Advance the fast weights and FIFO once
    \EndFor
    \State Save the last valid write's inputs for the next segment
    \State Accumulate the weighted segment gradient
\EndFor
\State Clip the accumulated gradient; take one optimizer and scheduler step
\end{algorithmic}
\smallskip\hrule
\end{minipage}\par\smallskip

\paragraph{Example at the first boundary}
Number the sampled decisions from zero. The first segment processes $0,\ldots,11$, and the second starts at 12. The loss at decision 12 can act directly on the replayed entry for 11 and on its writer. The older entries at 7--10 can still inform decision 12, but its gradient does not traverse the operations that originally wrote those four entries. Value retention and the horizon for assigning credit are therefore different properties.

\paragraph{Which decisions train the outer model?}
Padding does not advance memory. A valid observation can advance memory even when its action target is not selected for supervision. Within a window, let $n=\sum_j n_j$. A segment with $n_j>0$ contributes its mean supervised loss with weight $n_j/n$ before gradient accumulation. For 63 supervised decisions, for example, the last segment contributes $3/63$, not the same weight as a 12-decision segment. The inner updates still follow the observation sequence. The training procedure above gives the window-level schedule, omitting distributed bookkeeping.

\paragraph{Noise and flow time}
Multiple noisy action--wrench targets share the observation prefix and the same wrench history. The joint data processor chooses the number of flow samples to fit the available sequence budget. For an unconstrained noisy block it draws $u\sim\mathrm{Beta}(1.5,1)$ and sets $\tau=1-u$: the noise endpoint is $\tau=0$ and the clean endpoint is $\tau=1$. These are additional denoising examples for the same observation, not additional temporal decisions or writes to the bank.

\paragraph{Reproducing a boundary write}
The forward and backward requirements can be checked separately. With slow weights held fixed, compare the replayed state to the original post-write state: the entry timestamps and FIFO order should agree exactly, and the feature values and fast-state increments should agree up to numerical precision. For the backward check, use the first supervised prediction that reads the replayed entry. Its loss has a path to the writer and the learned initialization, but no path to the computation that produced the detached pre-write state. Matching the FIFO contents alone tests the forward requirement only.

\clearpage
\section{Policy Forward Pass}
\label{app:policy_forward}

\paragraph{Matching the training time axis}
At deployment, the robot executes the first 16 actions of a 32-step forecast before the next decision, matching the training stride. The front-camera timestamp anchors the observation: we select the nearest wrist-camera frames and interpolate proprioception to that time. At decision $t$ the policy reads only $B_{t-1}$ and processes the current views independently. It commits $E_t$ once, making it available at $t+1$ (Algorithm~\ref{alg:forward}). The bank starts empty; retries reuse the cached prediction without a second write. Memory resets at the start of a rollout, not at a training-window or TBPTT boundary.

\begin{figure}[H]
\centering
\includegraphics[width=\columnwidth]{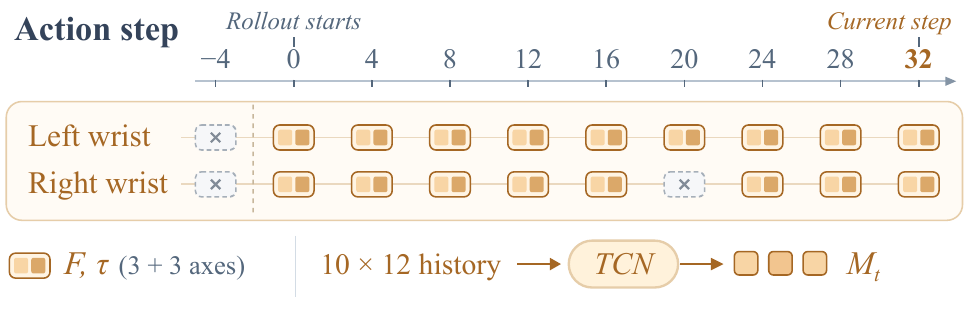}
\caption{\textbf{Sampling the two wrist wrenches.} At $n=32$, step $-4$ precedes the rollout. At step 20 only the right wrist is masked in this example.}
\label{fig:supp_wrench_sampling}
\end{figure}

\paragraph{Matching wrench history}
At deployment we retain the training normalization and disable image augmentation. At action step $n$, we sample both wrists at $n-36,n-32,\ldots,n$ (Fig.~\ref{fig:supp_wrench_sampling}). Each of the ten samples contains three force and three torque components per wrist in the world frame. We zero and mask pre-rollout slots and missing wrist readings without copying later samples.

\subsection{Building the policy inputs at a decision}

\paragraph{Addressing the visual memories}
The bank is passed to the VLM before the three current views. Within a bank entry the three camera spans retain their camera labels and their original patch grids for visual rotary position encoding. We reuse the stored features; the image encoder does not process the historical RGB images again at inference. To identify the age of an entry written at frame $f_s$ and read at $f_t$, its prefix records $f_s-f_t$, rather than just its FIFO slot number. The current views keep their own image spans. Consequently an empty bank at the start of a rollout does not require substituting copies of the first frame.

\begin{figure}[H]
\centering
\includegraphics[width=\columnwidth]{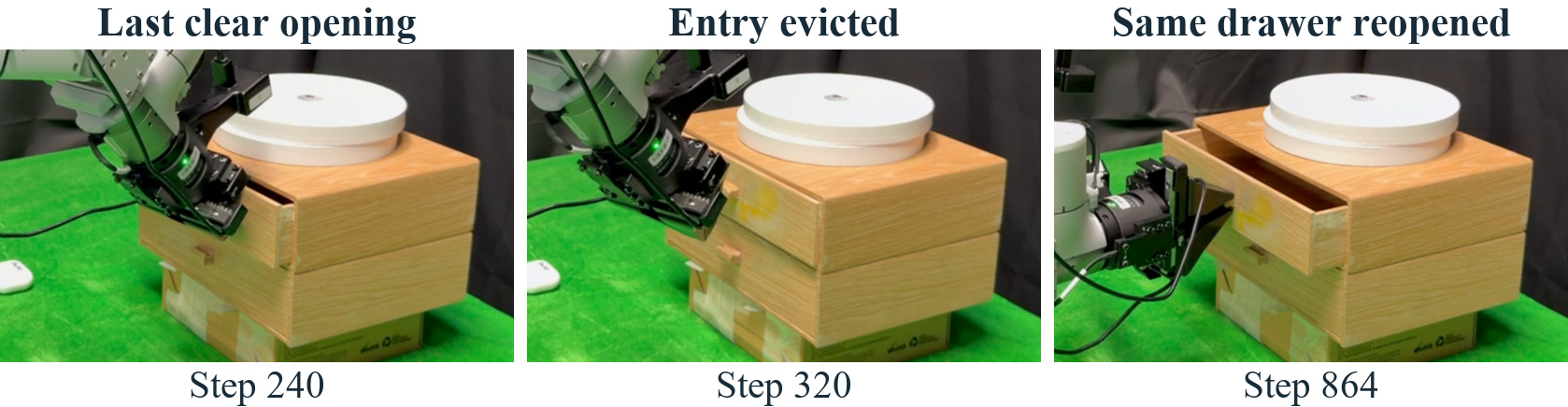}
\caption{\textbf{Initial-scene recall beyond the explicit window.} The external camera shows the last clearly visible opening at a sampled decision (31~s), the scene when its entry is evicted (41~s), and the same drawer reopening (105~s). Video times are approximate. We align the arm's motion and pauses with the logged action times.}
\label{fig:supp_drawer_trace}
\end{figure}

\paragraph{Recall after a visual entry leaves the bank}
Figure~\ref{fig:supp_drawer_trace} follows one OpenDrawerMem rollout. At action step 240, the upper drawer's interior is still clearly visible in the external view. The robot then closes the drawer. The bank log records five further writes at steps 256, 272, 288, 304, and 320. The write at 320 evicts entry 240. By step 864, the robot has reopened the same drawer. At that decision, the five entries available to the policy are from steps 784 through 848.

\paragraph{Preserving the order of contact events}
The wrench encoder projects each 12-coordinate row to 256 channels. Three residual temporal convolutions with kernel width three and dilations $1,2,4$ use left padding only. The output has ten rows; there is no pooling over time. An encoded row can therefore summarize preceding contact changes without incorporating a later measurement. The coordinate mask is applied before the projection. A row with at least one valid wrist remains available; a wholly missing row is zeroed after each residual block and excluded from attention.

\paragraph{Using the encoded history}
At each full-attention layer, the incoming denoising features supply the queries and the ten encoded rows supply the keys and values of an eight-head wrench cross-attention. The context gate takes the VLM gate-token readout, all ten ordered wrench features, and their coordinate-validity flags as inputs. It produces one scalar for that layer and observation, shared across the denoising tokens. Its linear projection is initialized to zero, so the wrench residual starts at zero; after training, its contribution can vary with both task context and contact history. When the entire history is missing, the residual is zero irrespective of the gate. This is why an absent measurement and a measured zero wrench are treated differently.

\begin{algorithm}[H]
\caption{M\textsuperscript{2}AdaptVLA forward pass at decision $t$}
\label{alg:forward}
\small
\begin{algorithmic}[1]
\Require Instruction $\ell$, views $I_t$, state $p_t$, recent wrenches $w_{\le t}$, and previous memory $(F_{t-1},B_{t-1})$
\State $Z_t \gets E_{\rm vis}(I_t)$, \quad $M_t \gets \operatorname{TCN}(w_{\le t})$
\State $\mathcal C_t \gets (\ell,p_t,Z_t,B_{t-1})$
\State Initialize joint action--wrench trajectory $Y$ from noise
\For{$k=1,\ldots,S$}
    \State Predict the joint velocity with $v_\theta(Y,\tau;\mathcal C_t,M_t)$,
    \Statex \hspace{\algorithmicindent}using the context gate to scale the wrench cross-attention output at full-attention layers (Eq.~\eqref{eq:wrench_fusion})
    \State Advance $Y$ along the predicted velocity
\EndFor
\State $(A_t,W_t^+) \gets \operatorname{split}(Y)$
\State Update $F_{t-1}$ with $Z_t$ to obtain $F_t$ by Eq.~\eqref{eq:ttt_update}
\State Read $R_t$ from $F_t$ and fuse it with $Z_t$ to form $E_t$
\State Append $E_t$ to the bounded bank by Eq.~\eqref{eq:bank_update}
\State \Return $(A_t,W_t^+)$ and next memory $(F_t,B_t)$
\end{algorithmic}
\end{algorithm}

\clearpage
\section{Wrench Feedback for Online Adaptation}
\label{app:wrench_feedback}

The mask $\Omega_t$ in Eq.~\eqref{eq:adaptation_loss} pairs a wrench forecast with the measurement available when the corresponding action finishes. In the whiteboard experiments, we sample executed steps 4, 8, 12, and 16 of the same predicted chunk. At each step we assess the two wrists independently. A wrist contributes all six force and torque coordinates if the complete sample is finite, was received by step completion, and has a sensor timestamp within the preceding 8\,ms. This timing rule keeps stale readings from supervising a later action. We skip the update when no coordinates qualify.

\paragraph{Comparing forecasts and measurements in the same units}
For the whiteboard policy we transform each valid raw wrench coordinate $w_d$ using fixed training statistics. Let $q_d^-$ and $q_d^+$ be its training quantiles at probabilities 0.001 and 0.999, and put $s_d=\max(q_d^+-q_d^-,10^{-6})$. The normalized measurement is
\begin{equation}
 \bar w_d=\operatorname{clip}_{[-1,1]}
       \!\left(2\frac{w_d-q_d^-}{s_d}-1\right).
 \label{eq:supp_wrench_norm}
\end{equation}
We use this transform for input histories, training targets and online feedback. Force and torque coordinates therefore enter the loss in the same normalized units as their respective predictions. A missing coordinate is masked, not interpreted as an observed force of zero. At most $4\times2\times6=48$ coordinates can supervise one update. If a wrist is unavailable at one eligible step, the other wrist at that step and the measurements at other eligible steps can still contribute.

\paragraph{The online gradient path}
Let $U=\operatorname{sg}(Y^{(9)})$ denote the state after the ninth Euler step and $C$ the cached conditioning inputs of the original decision. Under adapters $\phi$ we reconstruct the final forecast as
\begin{equation}
 \widehat W^+(\phi)=\Pi_W\!\left[ U+\Delta\tau\,
             v_{\theta,\phi}(U,\tau_9;C)\right],
 \label{eq:supp_last_step}
\end{equation}
where $\Pi_W$ selects the wrench coordinates. Equation~\eqref{eq:adaptation_loss} compares the selected output rows to Eq.~\eqref{eq:supp_wrench_norm}. Its gradient reaches the adapters through this last evaluation of the shared model. No tape for the preceding nine denoising steps is required. The joint action output is not treated as an expert label or a target during replay.

\begin{figure}[!t]
    \centering
    \includegraphics[width=\columnwidth]{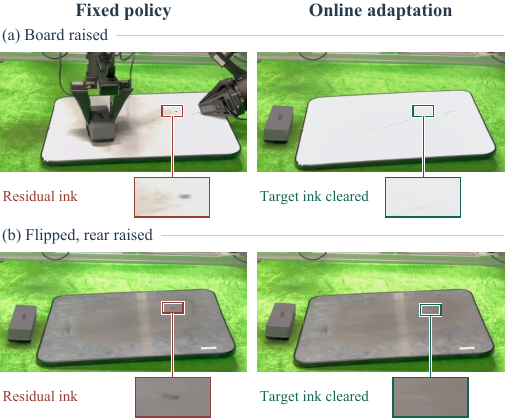}
    \caption{\textbf{Wiping under two contact shifts.} The entire board is raised (top), or the board is flipped and its rear edge is raised (bottom). In both examples, a short ink mark remains with the fixed policy (left); predictive test-time adaptation clears the corresponding target region (right). Insets enlarge the boxed regions.}
    \label{fig:supp_wiping_outcomes}
\end{figure}

\paragraph{Adapter scope and timing}
We deploy the whiteboard checkpoint from 10,000 offline updates. Replay does not modify the visual fast weights or the FIFO.

We insert rank-16 LoRA adapters ($\alpha_{\rm LoRA}=16$, so $\alpha_{\rm LoRA}/r=1$) in layers 1--4 of both the VLM and the action expert. We update them with AdamW at $10^{-5}$, zero weight decay, and a gradient norm limit of 1.0. Base weights stay fixed, and the adapters and optimizer state reset between rollouts. The fixed control uses the same checkpoint and both memories with the adapter update disabled (Fig.~\ref{fig:supp_wiping_outcomes}).

The latency in Table~\ref{tab:wipe_adaptation} measures the update itself, excluding sensor collection, action execution, and generation of the next action chunk. We time the two adapter configurations sequentially on one RTX~5090, discard five warmup updates, and average the next 20 for each configuration. Table~\ref{tab:supp_update_distribution} reports the timing distribution for the same measurements, including the 95th percentile.

\begin{table}[H]
\small\centering
\caption{Server-side update time (ms) on one RTX~5090; 20 timed updates per configuration. The fixed policy makes no update.}
\label{tab:supp_update_distribution}
\setlength{\tabcolsep}{4pt}
\begin{tabular*}{\columnwidth}{@{\extracolsep{\fill}}lrrr@{}}
\toprule
Adapted layers & Mean & Median & 95th percentile \\
\midrule
1--4  & 159.5 & 157.9 & 167.4 \\
1--24 & 267.7 & 245.4 & 282.5 \\
\bottomrule
\end{tabular*}
\end{table}

\clearpage
\twocolumn[{
\section{Online Adaptation for Raised Plug Insertion}
\label{app:raised_plug}
\noindent
We supplement the whiteboard evaluation with an additional online adaptation experiment on plug insertion under changed contact conditions.
\par\medskip
\begingroup\centering
\includegraphics[width=\textwidth]{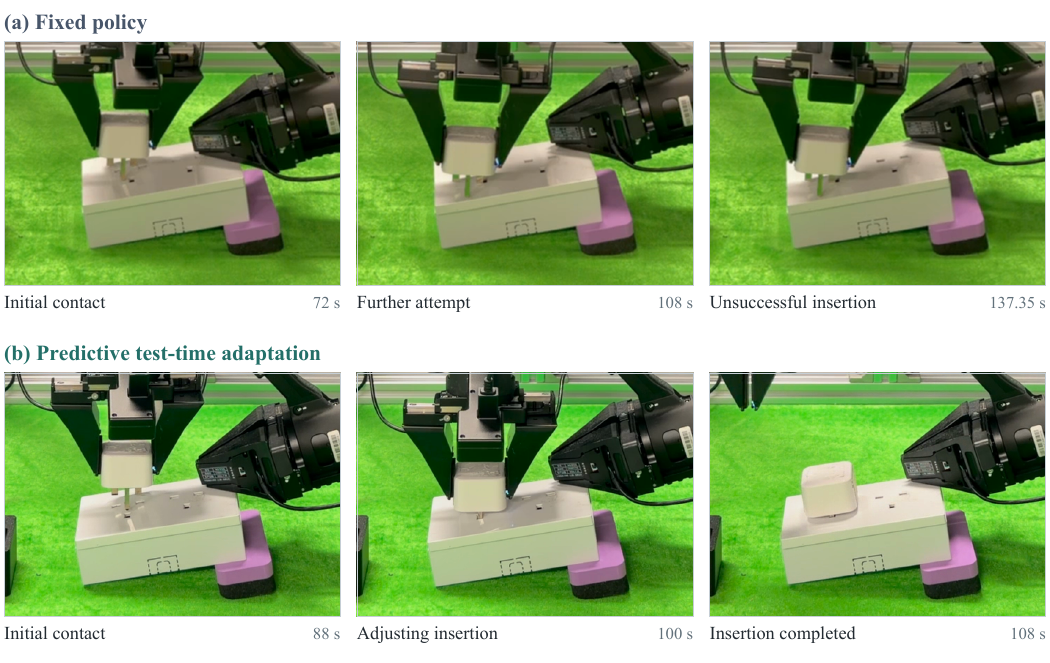}
\captionsetup{hypcap=false}
\captionof{figure}{\textbf{Plug insertion with a raised power strip.} A board eraser raises one side of the strip. In the fixed-policy example (top), repeated attempts leave the plug unseated. With predictive test-time adaptation using LoRA in layers 1--4 (bottom), the robot adjusts its insertion and releases the seated plug. The rows show separate illustrative trials; timestamps refer to their respective recordings.}
\label{fig:supp_raised_plug}
\par\endgroup
\vspace{\textfloatsep}
}]

\paragraph{Setup}
A board eraser placed beneath one side of the power strip changes the height and orientation of its sockets. We compare the fixed policy with online adaptation using LoRA in layers 1--4 or 1--24, with 20 trials per configuration. Success requires fully inserting the plug and releasing it.

\begin{table}[H]
\small\centering
\caption{\textbf{Predictive test-time adaptation on Insert Plug.} Success rates are based on 20 trials per policy. Parentheses give the relative change from the fixed policy.}
\label{tab:supp_raised_plug}
\setlength{\tabcolsep}{2pt}
\begin{tabular*}{\columnwidth}{@{\extracolsep{\fill}}lrrr@{}}
\toprule
 & Fixed & LoRA 1--4 & LoRA 1--24 \\
\midrule
Strip raised (\%) & \underline{75} & \textbf{85}\,{\scriptsize $(+13.3\%)$} & 70\,{\scriptsize $(-6.7\%)$} \\
\bottomrule
\end{tabular*}
\end{table}

\newpage
\paragraph{Results}
As shown in Table~\ref{tab:supp_raised_plug}, adapting layers 1--4 raises success from 75\% to 85\%, a relative improvement of 13.3\%, whereas adapting all 24 layers achieves 70\%. Consistent with the whiteboard results, adapting more layers does not necessarily improve robustness to changed contact conditions. Figure~\ref{fig:supp_raised_plug} illustrates the benefit of the four-layer configuration: the fixed policy repeatedly attempts insertion without seating the plug, whereas the adaptive policy completes insertion after an initially unsuccessful attempt. These results extend the evidence for predictive adaptation to contact changes in plug insertion.

%% file: main.bib
@inproceedings{memoryvla,
  title = {{MemoryVLA}: Perceptual-Cognitive Memory in Vision-Language-Action Models for Robotic Manipulation},
  author = {Shi, Hao and Xie, Bin and Liu, Yingfei and Sun, Lin and Liu, Fengrong and Wang, Tiancai and Zhou, Erjin and Fan, Haoqiang and Zhang, Xiangyu and Huang, Gao},
  booktitle = {International Conference on Learning Representations},
  year = {2026},
  url = {https://arxiv.org/abs/2508.19236}
}

@misc{fmvla,
  title = {{FM-VLA}: Force-based Memory for Vision-Language-Action Models in Contact-Rich Manipulation},
  author = {Li, Ruicheng and Li, Qixiu and Ma, Ruichun and Deng, Yu and Luo, Lin and Du, Zhiying and Xiang, Jianfeng and Liang, Huizhi and Wang, Ruicheng and Yang, Jiaolong and Guo, Baining},
  year = {2026},
  eprint = {2607.18231},
  archivePrefix = {arXiv},
  primaryClass = {cs.RO},
  note = {Accepted at CoRL 2026},
  url = {https://arxiv.org/abs/2607.18231}
}

@inproceedings{ttt,
  title = {Learning to ({Learn at Test Time}): {RNNs} with Expressive Hidden States},
  author = {Sun, Yu and Li, Xinhao and Dalal, Karan and Xu, Jiarui and Vikram, Arjun and Zhang, Genghan and Dubois, Yann and Chen, Xinlei and Wang, Xiaolong and Koyejo, Sanmi and Hashimoto, Tatsunori and Guestrin, Carlos},
  booktitle = {Proceedings of the 42nd International Conference on Machine Learning},
  year = {2025},
  series = {Proceedings of Machine Learning Research},
  volume = {267},
  pages = {57503--57522},
  publisher = {PMLR},
  url = {https://proceedings.mlr.press/v267/sun25h.html}
}

@misc{t2mem,
  title = {{$T^2$Mem}: Learning Test-Time Memory for Robotics},
  author = {Liu, Yize and Huang, Huang and Hong, Yining and Du, Zijian and Cao, Zhi and Fei-Fei, Li and Wu, Jiajun},
  year = {2026},
  eprint = {2609.36720},
  archivePrefix = {arXiv},
  primaryClass = {cs.RO},
  url = {https://arxiv.org/abs/2609.36720}
}

@misc{tttrm,
  title = {Test-Time Training as Residual Memory for Robot Policies},
  author = {Wang, Haoxuan and Zhang, Gengyu and Kompella, Ramana Rao and Liu, Gaowen and Yan, Yan},
  year = {2026},
  eprint = {2610.04701},
  archivePrefix = {arXiv},
  primaryClass = {cs.RO},
  url = {https://arxiv.org/abs/2610.04701}
}

@misc{adaworldpolicy,
  title = {{AdaWorldPolicy}: World-Model-Driven Diffusion Policy with Online Adaptive Learning for Robotic Manipulation},
  author = {Yuan, Ge and Qiao, Qiyuan and Zhang, Jing and Xu, Dong},
  year = {2026},
  eprint = {2602.20057},
  archivePrefix = {arXiv},
  primaryClass = {cs.RO},
  url = {https://arxiv.org/abs/2602.20057}
}

@inproceedings{lora,
  title = {{LoRA}: Low-Rank Adaptation of Large Language Models},
  author = {Hu, Edward J. and Shen, Yelong and Wallis, Phillip and Allen-Zhu, Zeyuan and Li, Yuanzhi and Wang, Shean and Wang, Lu and Chen, Weizhu},
  booktitle = {International Conference on Learning Representations},
  year = {2022},
  url = {https://openreview.net/forum?id=nZeVKeeFYf9}
}

@misc{damvla,
  title = {{DAM-VLA}: Decoupled Asynchronous Multimodal Vision Language Action model},
  author = {Vanjani, Pankhuri and Li, Zhuoyue and Suliga, Jakub and Reuss, Moritz and Geraci, Gianluca and Jiang, Xinkai and Lioutikov, Rudolf},
  year = {2026},
  eprint = {2606.12105},
  archivePrefix = {arXiv},
  primaryClass = {cs.RO},
  note = {Accepted at CoRL 2026},
  url = {https://arxiv.org/abs/2606.12105}
}

@inproceedings{forcevla,
  title = {{ForceVLA}: Enhancing VLA Models with a Force-aware MoE for Contact-rich Manipulation},
  author = {Yu, Jiawen and Liu, Hairuo and Yu, Qiaojun and Ren, Jieji and Hao, Ce and Ding, Haitong and Huang, Guangyu and Huang, Guofan and Song, Yan and Cai, Panpan and Zhang, Wenqiang and Lu, Cewu},
  booktitle = {Advances in Neural Information Processing Systems},
  volume = {38},
  year = {2025},
  doi = {10.52202/085713-3124},
  url = {https://papers.nips.cc/paper_files/paper/2025/hash/8633b46e12cc5f2ee1f05a6ca2c65b38-Abstract-Conference.html}
}

@inproceedings{sam2act,
  title = {{SAM2Act}: Integrating Visual Foundation Model with A Memory Architecture for Robotic Manipulation},
  author = {Fang, Haoquan and Grotz, Markus and Pumacay, Wilbert and Wang, Yi Ru and Fox, Dieter and Krishna, Ranjay and Duan, Jiafei},
  booktitle = {Proceedings of the 42nd International Conference on Machine Learning},
  pages = {15925--15942},
  year = {2025},
  volume = {267},
  series = {Proceedings of Machine Learning Research},
  publisher = {PMLR},
  url = {https://proceedings.mlr.press/v267/fang25c.html}
}

@inproceedings{memer,
  title = {{MemER}: Scaling Up Memory for Robot Control via Experience Retrieval},
  author = {Sridhar, Ajay and Pan, Jennifer and Sharma, Satvik and Finn, Chelsea},
  booktitle = {International Conference on Learning Representations},
  year = {2026},
  url = {https://arxiv.org/abs/2510.20328}
}

@misc{mem,
  title = {{MEM}: Multi-Scale Embodied Memory for Vision Language Action Models},
  author = {Torne, Marcel and Pertsch, Karl and Walke, Homer and Vedder, Kyle and Nair, Suraj and Ichter, Brian and Ren, Allen Z. and Wang, Haohuan and Tang, Jiaming and Stachowicz, Kyle and Dhabalia, Karan and Equi, Michael and Vuong, Quan and Springenberg, Jost Tobias and Levine, Sergey and Finn, Chelsea and Driess, Danny},
  year = {2026},
  eprint = {2603.03596},
  archivePrefix = {arXiv},
  primaryClass = {cs.RO},
  note = {Accepted at CoRL 2026 (Spotlight)},
  url = {https://arxiv.org/abs/2603.03596}
}

@misc{robottt,
  title = {{RoboTTT}: Context Scaling for Robot Policies},
  author = {Jiang, Yunfan and Chebotar, Yevgen and Zheng, Ruijie and Hu, Fengyuan and Ge, Yunhao and Wu, Jimmy and Dai, Tianyuan and Reed, Scott and Fei-Fei, Li and Zhu, Yuke and Fan, Linxi},
  year = {2026},
  eprint = {2607.15275},
  archivePrefix = {arXiv},
  primaryClass = {cs.RO},
  url = {https://arxiv.org/abs/2607.15275}
}

@misc{salt,
  title = {Adapting Vision-Language-Action Models to Unknown Visual Disruptions During Execution},
  author = {Lee, Ahin and Seo, Jinwoo and Jang, Youngsoo and Gong, Taesik},
  year = {2026},
  eprint = {2610.07946},
  archivePrefix = {arXiv},
  primaryClass = {cs.RO},
  url = {https://arxiv.org/abs/2610.07946}
}

@misc{tavla,
  title = {{TA-VLA}: Elucidating the Design Space of Torque-aware Vision-Language-Action Models},
  author = {Zhang, Zongzheng and Xu, Haobo and Yang, Zhuo and Yue, Chenghao and Lin, Zehao and Gao, Huan-ang and Wang, Ziwei and Zhao, Hao},
  year = {2025},
  eprint = {2509.07962},
  archivePrefix = {arXiv},
  primaryClass = {cs.RO},
  note = {Accepted at CoRL 2025},
  url = {https://arxiv.org/abs/2509.07962}
}

@article{memoact,
  title = {{MemoAct}: At\-kin\-son--Shiffrin-Inspired Hierarchical Memory-Augmented Policy for Robotic Manipulation},
  author = {Tan, Liufan and Li, Jiale and Jing, Gangshan},
  journal = {IEEE Robotics and Automation Letters},
  year = {2026},
  volume = {11},
  number = {9},
  pages = {10927--10934},
  doi = {10.1109/LRA.2026.3719233},
  url = {https://doi.org/10.1109/LRA.2026.3719233}
}

@misc{rememvla,
  title = {{ReMem-VLA}: Empowering Vision-Language-Action Model with Memory via Dual-Level Recurrent Queries},
  author = {Li, Hang and Shen, Fengyi and Chen, Dong and Yang, Liudi and Wang, Xudong and Shi, Jinkui and Bing, Zhenshan and Liu, Ziyuan and Knoll, Alois},
  year = {2026},
  eprint = {2603.12942},
  archivePrefix = {arXiv},
  primaryClass = {cs.RO},
  url = {https://arxiv.org/abs/2603.12942}
}

@inproceedings{tracevla,
  title = {{TraceVLA}: Visual Trace Prompting Enhances Spatial-Temporal Awareness for Generalist Robotic Policies},
  author = {Zheng, Ruijie and Liang, Yongyuan and Huang, Shuaiyi and Gao, Jianfeng and Daum{\'e} III, Hal and Kolobov, Andrey and Huang, Furong and Yang, Jianwei},
  booktitle = {International Conference on Learning Representations},
  year = {2025},
  url = {https://proceedings.iclr.cc/paper_files/paper/2025/hash/8667f264f88c7938a73a53ab01eb1327-Abstract-Conference.html}
}

@misc{contextvla,
  title = {{ContextVLA}: Vision-Language-Action Model with Amortized Multi-Frame Context},
  author = {Jang, Huiwon and Yu, Sihyun and Kwon, Heeseung and Jeon, Hojin and Seo, Younggyo and Shin, Jinwoo},
  year = {2025},
  eprint = {2510.04246},
  archivePrefix = {arXiv},
  primaryClass = {cs.RO},
  url = {https://arxiv.org/abs/2510.04246}
}

@article{wallx,
  title = {Igniting {VLMs} toward the Embodied Space},
  author = {Zhai, Andy and Liu, Brae and Fang, Bruno and Cai, Chalse and Ma, Ellie and Yin, Ethan and Wang, Hao and Zhou, Hugo and Wang, James and Shi, Lights and Liang, Lucy and Wang, Make and Wang, Qian and Gan, Roy and Yu, Ryan and Li, Shalfun and Liu, Starrick and Chen, Sylas and Chen, Vincent and Xu, Zach},
  journal = {arXiv preprint arXiv:2509.11766},
  year = {2025},
  url = {https://arxiv.org/abs/2509.11766}
}

@misc{qwen35,
  author = {{Qwen Team}},
  title = {{Qwen3.5}: Towards Native Multimodal Agents},
  month = feb,
  year = {2026},
  url = {https://qwen.ai/blog?id=qwen3.5}
}

@inproceedings{gateddeltanet,
  author = {Yang, Songlin and Kautz, Jan and Hatamizadeh, Ali},
  title = {Gated Delta Networks: Improving {Mamba2} with Delta Rule},
  booktitle = {The Thirteenth International Conference on Learning Representations},
  year = {2025},
  url = {https://openreview.net/forum?id=r8H7xhYPwz}
}

@misc{frvla,
  author = {He, Jiaju and Yang, Zhiyuan and Han, Ruyang and Tang, Jiahong and Lin, Zhiping and Zhu, Haiyue},
  title = {{FR-VLA}: Implicit Force Reasoning {VLA} for Contact-Rich Manipulation},
  year = {2026},
  eprint = {2610.25981},
  archivePrefix = {arXiv},
  primaryClass = {cs.RO},
  url = {https://arxiv.org/abs/2610.25981}
}

@inproceedings{pi05,
  author = {Black, Kevin and Brown, Noah and Darpinian, James and Dhabalia, Karan and Driess, Danny and Esmail, Adnan and others},
  title = {{$\pi_{0.5}$}: A Vision-Language-Action Model with Open-World Generalization},
  booktitle = {Proceedings of the 9th Conference on Robot Learning},
  series = {Proceedings of Machine Learning Research},
  volume = {305},
  pages = {17--40},
  publisher = {PMLR},
  year = {2025},
  url = {https://proceedings.mlr.press/v305/black25a.html}
}

@inproceedings{pad,
  author = {Hansen, Nicklas and Jangir, Rishabh and Sun, Yu and Aleny{\`a}, Guillem and Abbeel, Pieter and Efros, Alexei A. and Pinto, Lerrel and Wang, Xiaolong},
  title = {Self-Supervised Policy Adaptation during Deployment},
  booktitle = {International Conference on Learning Representations},
  year = {2021},
  url = {https://arxiv.org/abs/2007.04309}
}

@inproceedings{scout,
  author = {Li, Yishu and Geng, Liyuan and Mao, Xinyi and Li, Amber and Held, David},
  title = {Scouting the Dynamics Gap: Test-Time Policy Adaptation via Action-Outcome Feedback},
  booktitle = {Conference on Robot Learning},
  year = {2026},
  url = {https://arxiv.org/abs/2609.36107}
}

@inproceedings{robomme,
  title = {{RoboMME}: Benchmarking and Understanding Memory for Robotic Generalist Policies},
  author = {Dai, Yinpei and Fu, Hongze and Lee, Jayjun and Liu, Yuejiang and Zhang, Haoran and Yang, Jianing and Finn, Chelsea and Fazeli, Nima and Chai, Joyce},
  booktitle = {Proceedings of the 43rd International Conference on Machine Learning},
  series = {Proceedings of Machine Learning Research},
  volume = {306},
  pages = {22396--22450},
  year = {2026},
  url = {https://proceedings.mlr.press/v306/dai26c.html}
}

@inproceedings{lee2019visiontouch,
  title = {Making Sense of Vision and Touch: Self-Supervised Learning of Multimodal Representations for Contact-Rich Tasks},
  author = {Lee, Michelle A. and Zhu, Yuke and Srinivasan, Krishnan and Shah, Parth and Savarese, Silvio and Li, Fei-Fei and Garg, Animesh and Bohg, Jeannette},
  booktitle = {International Conference on Robotics and Automation},
  pages = {8943--8950},
  year = {2019},
  doi = {10.1109/ICRA.2019.8793485}
}

@article{foar,
  title = {{FoAR}: Force-Aware Reactive Policy for Contact-Rich Robotic Manipulation},
  author = {He, Zihao and Fang, Hongjie and Chen, Jingjing and Fang, Hao-Shu and Lu, Cewu},
  journal = {IEEE Robotics and Automation Letters},
  volume = {10},
  number = {6},
  pages = {5625--5632},
  year = {2025},
  doi = {10.1109/LRA.2025.3560871}
}

@inproceedings{hora,
  title = {In-Hand Object Rotation via Rapid Motor Adaptation},
  author = {Qi, Haozhi and Kumar, Ashish and Calandra, Roberto and Ma, Yi and Malik, Jitendra},
  booktitle = {Proceedings of the 6th Conference on Robot Learning},
  series = {Proceedings of Machine Learning Research},
  volume = {205},
  pages = {1722--1732},
  year = {2023},
  url = {https://proceedings.mlr.press/v205/qi23a.html}
}

@misc{latentphysical,
  title = {Online Adaptation to Latent Physical Properties for Learning from Demonstration},
  author = {Kajitsuka, Kotaro and Murooka, Masaki and Kanehiro, Fumio},
  year = {2026},
  eprint = {2611.06118},
  archivePrefix = {arXiv},
  url = {https://arxiv.org/abs/2611.06118}
}

@article{vlamemorysurvey,
  author = {Wang, Yong and Shen, Boyu and Zhang, Xinwei and Li, Yizheng and Luo, Keyuan},
  title = {From Sliding Windows to Self-evolving Lifelong Memory: A Dual-perspective Survey of Memory Paradigms for Long-Horizon Robotics {VLA} Models},
  journal = {arXiv preprint arXiv:2610.23319},
  year = {2026},
  url = {https://arxiv.org/abs/2610.23319}
}

@inproceedings{decal,
  author = {Fu, Yankai and Chen, Ning and Zhao, Junkai and Zhang, Heng and Yao, Guocai and Wang, Pengwei and Wang, Zhongyuan and Zhang, Shanghang},
  title = {{DeCAL}: Towards Physically-Grounded Dexterous Vision-Language-Action Models via Contact-Aware Latent Co-Imagination},
  booktitle = {Conference on Robot Learning},
  year = {2026},
  url = {https://arxiv.org/abs/2609.09119}
}

@misc{eventvla,
  title = {{EventVLA}: Event-Driven Visual Evidence Memory for Long-Horizon Vision-Language-Action Policies},
  author = {Yang, Ganlin and Tu, Zhangzheng and Yang, Yuqiang and Mao, Sitong and Dong, Junyi and Chen, Tianxing and Peng, Jiaqi and Xiong, Jing and Cao, Jiafei and Dai, Jifeng and Zhou, Wengang and Mu, Yao and Wang, Tai},
  year = {2026},
  eprint = {2606.20092},
  archivePrefix = {arXiv},
  primaryClass = {cs.CV},
  url = {https://arxiv.org/abs/2606.20092}
}

@misc{kemo,
  title = {{KEMO}: Event-Driven Keyframe Memory for Long-Horizon Robot Manipulation with {VLA} Policies},
  author = {Zeng, Yihan and Ye, Minghao and Chen, Yiyuan and Shentu, Yide and Wu, Philipp and Yan, Zike and Li, Zhongyu},
  year = {2026},
  eprint = {2606.23589},
  archivePrefix = {arXiv},
  primaryClass = {cs.RO},
  url = {https://arxiv.org/abs/2606.23589}
}

@misc{replanmem,
  title = {When Plans Change, Remember: Replanning-Guided Visual Memory for Vision-Language-Action Models},
  author = {Smelova, Victoria and Kazachkov, Daniil and Kazantsev, Georgiy and Tsvetkova, Iuliia and Cherepanov, Egor and Kachaev, Nikita and Latyshev, Artem and Panov, Aleksandr I. and Kovalev, Alexey K.},
  year = {2026},
  eprint = {2610.22556},
  archivePrefix = {arXiv},
  primaryClass = {cs.RO},
  url = {https://arxiv.org/abs/2610.22556}
}

@inproceedings{guhurhistory,
  title = {Instruction-Driven History-Aware Policies for Robotic Manipulations},
  author = {Guhur, Pierre-Louis and Chen, Shizhe and Pinel, Ricardo Garcia and Tapaswi, Makarand and Laptev, Ivan and Schmid, Cordelia},
  booktitle = {Proceedings of the 6th Conference on Robot Learning},
  year = {2023},
  volume = {205},
  series = {Proceedings of Machine Learning Research},
  pages = {175--187},
  url = {https://proceedings.mlr.press/v205/guhur23a.html}
}

@inproceedings{fourdimvla,
  title = {{4D-VLA}: Spatiotemporal Vision-Language-Action Pretraining with Cross-Scene Calibration},
  author = {Zhang, Jiahui and Chen, Yurui and Xu, Yueming and Huang, Ze and Zhou, Yanpeng and Yuan, Yu-Jie and Cai, Xinyue and Huang, Guowei and Quan, Xingyue and Xu, Hang and Zhang, Li},
  booktitle = {Advances in Neural Information Processing Systems},
  year = {2025},
  volume = {38},
  doi = {10.52202/085713-1139}
}

@inproceedings{liberomem,
  title = {Rethinking Progression of Memory State in Robotic Manipulation: An Object-Centric Perspective},
  author = {Chung, Nhat and Hanyu, Taisei and Nguyen, Toan and Le, Huy and Bumgarner, Frederick and Nguyen, Duy Minh Ho and Vo, Khoa and Yamazaki, Kashu and Rainwater, Chase and Kieu, Tung and Nguyen, Anh and Le, Ngan},
  booktitle = {Proceedings of the AAAI Conference on Artificial Intelligence},
  volume = {40},
  pages = {3407--3415},
  year = {2026},
  doi = {10.1609/aaai.v40i5.37337}
}

@inproceedings{titans,
  title = {Titans: Learning to Memorize at Test Time},
  author = {Behrouz, Ali and Zhong, Peilin and Mirrokni, Vahab},
  booktitle = {Advances in Neural Information Processing Systems},
  year = {2025},
  volume = {38},
  doi = {10.52202/085713-3786}
}

@article{phaforce,
  title = {{PhaForce}: Phase-Scheduled Visual-Force Policy Learning with Slow Planning and Fast Correction for Contact-Rich Manipulation},
  author = {Wang, Mingxin and Yue, Zhirun and Lu, Renhao and Li, Yizhe and Wang, Zihan and Pan, Guoping and Dong, Kangkang and Cheng, Jun and Cheng, Yi and Liu, Houde},
  journal = {arXiv preprint arXiv:2603.08342},
  year = {2026},
  doi = {10.48550/arXiv.2603.08342}
}

@inproceedings{forcemimic,
  title = {{ForceMimic}: Force-Centric Imitation Learning with Force-Motion Capture System for Contact-Rich Manipulation},
  author = {Liu, Wenhai and Wang, Junbo and Wang, Yiming and Wang, Weiming and Lu, Cewu},
  booktitle = {IEEE International Conference on Robotics and Automation},
  year = {2025},
  doi = {10.1109/ICRA55743.2025.11128061}
}

@inproceedings{onlinesysid,
  title = {Preparing for the Unknown: Learning a Universal Policy with Online System Identification},
  author = {Yu, Wenhao and Tan, Jie and Liu, C. Karen and Turk, Greg},
  booktitle = {Robotics: Science and Systems XIII},
  year = {2017},
  doi = {10.15607/RSS.2017.XIII.048}
}

@inproceedings{memobench,
  title = {{MEMOBench}: A Process Level Memory Benchmark for Robotic Manipulation},
  author = {Sun, Haiyang and Wang, Haoxiao and Chen, Junming and Fang, Weicheng and Su, Zihao and Yi, Jingkun and Yi, Wenyou and Chen, Hao and Zhao, Zhou},
  booktitle = {Findings of the Association for Computational Linguistics: EMNLP 2026},
  year = {2026},
  pages = {7025--7040},
  url = {https://aclanthology.org/2026.findings-emnlp.333/}
}

@inproceedings{flowmatching,
  title = {Flow Matching for Generative Modeling},
  author = {Lipman, Yaron and Chen, Ricky T. Q. and Ben-Hamu, Heli and Nickel, Maximilian and Le, Matt},
  booktitle = {International Conference on Learning Representations},
  year = {2023},
  url = {https://openreview.net/forum?id=PqvMRDCJT9t}
}

@inproceedings{mikasa,
  author = {Cherepanov, Egor and Kachaev, Nikita and Kovalev, Alexey and Panov, Aleksandr},
  title = {Memory, Benchmark \& Robots: A Benchmark for Solving Complex Tasks with Reinforcement Learning},
  booktitle = {International Conference on Learning Representations},
  year = {2026},
  url = {https://openreview.net/forum?id=9cLPurIZMj}
}
